\documentclass{article}

\usepackage{arxiv}

\usepackage[T1]{fontenc}    
\usepackage[numbers]{natbib}

\usepackage{hyperref}       
\usepackage{url}            
\usepackage{booktabs}       
\usepackage{amsfonts}       
\usepackage{nicefrac}       
\usepackage{microtype}      
\usepackage{lipsum}
\usepackage{graphicx}
\graphicspath{ {./Images/} }

\usepackage{array}
\usepackage{multirow}
 \usepackage{amsmath} 
\usepackage{enumitem}         
\usepackage{booktabs}
\usepackage{nicefrac}
\usepackage{microtype}
\usepackage{float}

\title{Towards the Development of Balanced Synthetic Data for Correcting Grammatical Errors in Arabic: \\ An Approach Based on Error Tagging Model and Synthetic Data Generating Model}

\author{
  Ahlam Alrehili \\
  Department of Computer Sciences,\\ Faculty of Computing and Information Technology  \\
  King Abdulaziz University, Saudi Electronic University \\
  Medina, Saudi Arabia \\
  \texttt{a.alrehili@seu.edu.sa} 
  \And
  Areej Alhothali \\
  Department of Computer Sciences,\\ Faculty of Computing and Information Technology \\
  King Abdulaziz University \\
  Jeddah, Saudi Arabia \\
  \texttt{aalhothali@kau.edu.sa}
}

\begin{document}
\maketitle
\begin{abstract}
Synthetic data generation is widely recognized as a way to enhance the quality of neural grammatical error correction (GEC) systems. However, current approaches often lack diversity or are too simplistic to generate the wide range of grammatical errors made by humans, especially for low-resource languages such as Arabic. In this paper, we will develop the error tagging model and the synthetic data generation model to create a large synthetic dataset in Arabic for grammatical error correction. In the error tagging model, the correct sentence is categorized into multiple error types by using the DeBERTav3 model. Arabic Error Type Annotation tool (ARETA) is used to guide multi-label classification tasks in an error tagging model in which each sentence is classified into 26 error tags. The synthetic data generation model is a back-translation-based model that generates incorrect sentences by appending error tags before the correct sentence that was generated from the error tagging model using the ARAT5 model. In the QALB-14 and QALB-15 Test sets, the error tagging model achieved 94.42\% F1, which is state-of-the-art in identifying error tags in clean sentences. As a result of our syntactic data training in grammatical error correction, we achieved a new state-of-the-art result of F1-Score: 79.36\% in the QALB-14 Test set. We generate 30,219,310 synthetic sentence pairs by using a synthetic data generation model. Our data is accessible to the public\footnote{\url{https://github.com/Ahlam-alrehili/ArabGEC}}.
\end{abstract}


\section{Introduction}
Arabic grammar error correction (ArabGEC) has gained popularity in recent years because of the growing demand for robust and accurate natural language processing (NLP) tools for Arabic. A grammatical error correction system (GEC) corrects grammatical errors and other forms of error in writing automatically. It guarantees that written texts are coherent, concise, clear, well-organized, and effectively communicate the intended message. The GEC problem can be thought of as seq2seq (sequence-to-sequence) (converting incorrect sentences to grammatically correct sentences) and is inspired by neural machine translation (NMT)\cite{chollampatt2018multilayer}\cite{junczys2018approaching}. A correct and best choice will automatically be substituted for the incorrect ones in the context without modifying the syntax.

Among the official languages of the United Nations, Arabic is classified as a Semitic language, and there were 274 million speakers around the world\footnote{\url{https://www.statista.com/statistics/266808/the-most-spoken-languages-worldwide/}}. Natural language processing systems can find Arabic's grammatical structure challenging because of its complex grammatical structure. Moreover, the Arabic language has a rich morphology and a complex sentence structure\cite{shaalan2019challenges}. However, the lack of large, well-annotated datasets for Arabic grammatical error correction has hindered the development of robust and accurate models.

There are only two parallel corpora available for GEC research and they are QALB-14~\cite{mohit2014first} and QALB-15~\cite{rozovskaya2015second}. QALB-14 and QALB-15 are part of the Qatar Arabic Language Bank (QALB) project, which aims to compile a large collection of Arabic text that's been manually corrected, including news site comments, essays, and machine translations. A specialized annotation interface has been developed for this project, along with comprehensive annotation guidelines~\cite{jeblee2014cmuq}~\cite{obeid2013web}. A total of 20,430 and 1,542 samples are available in each of the two training corpora,(QALB-14) and (QALB-15). Despite the researchers' complete reliance on these data, it has some shortcomings, including inadequate coverage of Arabic language defects, and inconsistent punctuation correction, and is considered very few compared with other languages.

To overcome the limited availability of high-quality, annotated datasets, researchers have explored the use of synthetic data generation techniques, which involve the production of artificial instances of grammatical errors and their corresponding corrections. A synthetic data generation techniques technique increases the amount of training data available for grammatical error correction systems to be developed and evaluated. Various synthetic data generation techniques approaches, such as statistical machine translation and seq2seq models, have been investigated, with promising results in improving the performance of grammatical error correction models.

The generation of synthetic data for grammar correction has been extensively explored in many languages, but Arabic remains largely unexplored. Several articles have been published in the literature, but most of them use limited datasets and simplistic methodologies, usually using a rule-based approach. Even though Arabic has twice the lexicon size of English, there is little research dedicated to synthetic data in Arabic. Due to the lack of research on synthetic data generation for Arabic grammar correction, there is a significant gap in the linguistic and computational literature. Arabic's linguistic complexity and diversity make the lack of attention paid to synthetic data creation particularly noteworthy.

This paper aims to develop a methodology for producing balanced synthetic data that can be used to support the construction of a grammatical error correction model in the Arabic language. An error tagging model was designed based on the DeBERTav3 pre-train language model as a multi-label classification task so that we can predict the types of errors (tags) that may appear in correctly constructed sentences. Then, we combine the error tags with the correct sentences and create a model that generates synthetic parallel data called the synthetic data generation model. It uses the sequence-to-sequence technique with the AraT5 model to generate incorrect sentences that are consistent with the types of errors that were previously identified in the error tagging model. Using this model, synthetic errors are generated that are realistically replicating human errors, improving the performance of automated correction systems by giving the resulting sentences a realistic character. A qualitative contribution to the field of Arabic language processing is this methodology, which provides a rich and balanced set of data that helps improve grammatical correction models, thus improving their accuracy and effectiveness in processing and analyzing Arabic texts.


The remainder of this paper is organized as follows: The second section discusses techniques for generating synthetic
sentences in GEC. Section 3 describes the methodology used to build the GEC corpus. Section 4 presents an overview of Arabic synthetic data techniques, grammatical error correction multi-label classification, pre-trained BERT models, and large language Text-to-Text Transformers used in the literature will be provided. Section 4 presents the data used to train our models, and section 5 provides details of the methodology used. Section 6 presents the the evaluation metric and experimental setup and section 7 presents our experimental results. A discussion of our proposed methodology is presented in section 8, and section 9 summarizes our contributions and outlines future directions.

\section{Related Work}
In this section, we provide an overview of Arabic synthetic data techniques, grammatical error correction multi-label classification, pre-trained BERT models, and large language Text-to-Text Transformers used in literature.

\subsection{Arabic Synthetic Data Techniques}
This section discusses the synthetic data technique used in prior ArabGEC research. Solyman et al.~\cite{solyman2021synthetic} generated only spelling errors using the confusion function. The corrupted sentences were constructed by selecting random words in sentences, then deleting them, duplicating them, substituting letters within them, or inserting new letters. Every sentence had a probability distribution of 10\% for duplicating and deleting words and 40\% for changing letters or deleting characters. An Al-Watan 20021 \footnote{\url{https://sourceforge.net/projects/arabiccorpus/files/latest/download}}corpus was used. Originally, the corpus contained ten million words written by professional journalists in Modern Standard Arabic (MSA). Al-Watan newspaper provided the data, which is divided into six categories. In the first step, the data was compiled and reorganized into a single file, followed by the normalization of sentences containing emoticons, non-alphanumeric characters, and exclamation marks. After splitting the sentences into maximum lengths of 50 words, they dropped sentences that were less than 5 words in length. In training and development corpora, 18,061,610 million words were generated. Synthetic examples generated are of low quality and contain only spelling errors, making them inefficient.

Solyman et al.~\cite{solyman2022automatic}  developed a technique called the noising method for generating synthetic training data for ArabGEC models. It consists of a combination of back-translation and direct error injection. Despite using a neural noise model based on multi-head attention, back translation cannot predict sentences correctly because it is based on a limited dataset. A sequence of four words is randomly selected and fed into the back translation model for each original sentence of more than ten words. This technique generates spelling, syntax, and grammar errors without predicting their types and numbers. A direct error injection inserts spelling errors directly into words or normalizes Arabic characters that have a similar shape. They used the Open Source International Arabic News (OSIAN) corpus\footnote{\url{https://wortschatz.uni-leipzig.de/en/download/Arabic}} containing 15,001,707 sentences and 367,572,569 words. The data was preprocessed and cleaned of mentions, non-UTF8 encoding, diacritics, hashtags, over-space, and links. Additionally, they divided the text into sentences of at least five words each. After data cleaning, there were 13,333,929 sentences.

Solyman et al.~\cite{solyman2023optimizing} proposed data augmentation approaches for ArabGEC. They describe seven specific approaches to data augmentation: Misspelling, Swapping, Tokens, Sources, Reverses, Monos, and Replaces.  Misspelling involves inserting or deleting spelling errors within a target sentence. In Swap, a random swap is performed to create a new word order in the target sentence. Token involves substituting specific words for an alternative token in a target sentence, such as <UNK>. A source sentence is copied to a target sentence to improve accuracy. In reverse, the target sentence is rearranged to enhance the encoder's contribution. Mono: Aligns target-side words monotonously with the source, making the target sentences less fluent and encouraging the encoder's representations to take precedence. In the Replace method, random entries from the training vocabulary are used to replace the target words.  These techniques are implemented during training to improve the training data by improving target sentences, which are then used as auxiliary tasks to strengthen the encoder.  Using each approach, the augmented examples are appended to the original training data, increasing the diversity and size of the dataset for the GEC model to be trained. They used synthetic data from a previous study by~\cite{solyman2022automatic} to augment the data. The training and development sets consisted of 1,500,173 sentence pairs.

Kwon et al.~\cite{kwon2023chatgpt} discussed the use of synthetic data to improve the performance of AraGEC models. Three different methods of data augmentation are explored by the authors: First, they utilized ChatGPT as a corrupter to introduce grammatical errors into correct Arabic sentences, creating a parallel dataset. Using a taxonomy from the Arabic Learner Corpus, they ensure a variety of error types have been identified. The second type is token noise and error adaptation, which involves introducing random alterations to characters and words, such as adjusting spaces and punctuation. Matching error rates with the benchmark dataset is the goal. The third type is called reverse noise, where they use beam search to convert clean sentences into noisy ones. QALB-2014 and 2015 datasets are used for training one reverse model, and ChatGPT is used to generate a parallel dataset. The authors compared the performance of fine-tuned models on a training dataset with models on artificially created datasets to evaluate the efficacy of these methods. However, when fine-tuning on out-of-domain examples, the performance drops significantly when compared to those tuned on in-domain synthetic data. It emphasizes the importance of high-quality, relevant synthetic data for effective model training. Furthermore, the researchers found that as dataset size increases, precision and recall trade-off, with precision improving and recall declining. To fully optimize ArabGEC systems, data augmentation techniques, and synthetic data quality need to be improved, even though synthetic data generation shows promise.

 The lack of data makes Arabic natural language processing (NLP) challenging. QALB-14 and QALB-15 are the two data sets that make up the majority of ArabGEC (QALB-14, QALB-15). There are relatively few examples in shared data, that consist of punctuation errors, and do not cover all kinds of errors. Moreover,  ArabGEC has limited synthetic data generation capabilities and is relatively unexplored compared to other languages. The quality of augmented data is crucial to the effectiveness of the methodology. Performance could be negatively impacted if unrealistic or irrelevant patterns are introduced during the augmentation process. Only four researches have been conducted on the generation of Arabic synthetic data, as discussed in the preceding paragraphs.  Furthermore, most previous research focused on generating only spelling errors and neglected other types of errors. In addition, some research focused on generating errors at the word level, such as adding or replacing words. Despite this, Arabic is one of the Semitic languages and has a rich morphology, making existing synthetic data sufficient to improve the quality of ArabGEC. Thus, further research is needed to unlock the potential of ArabGEC.
 \subsection{Arabic Grammatical Error Correction as Multi-label Classification Task}
 In Arabic text, the Arabic Error Type Annotation (ARETA) tool~\cite{belkebir2021automatic}, a system for automatically annotating Modern Standard Arabic, several steps are required: First, align raw input text with reference text corrected versions
Second: automatically determines error types. It involves a four-part process:
\begin{itemize}[noitemsep, topsep=0pt]
\item Punctuation: Determines whether punctuation errors have occurred using regular expressions.
\item Regex: Identifies splits, mergers, deletions, insertions, and other orthographic errors using regular expressions.
\item Ortho-Morph: A morphological analyzer based on CAMeL Tools is used to handle complex orthographic and morphological errors. Possible edits are generated and the shortest paths to convert the raw word into the reference word are determined, which error tags are associated with these edits.
\item Multi-Word: Processes one-to-many word pairs by tokenizing them with Arabic Treebank and assigning errors to the tokens
\end{itemize}
Finally, ARETA can be used to evaluate grammatical error correction systems by comparing predicted error tags with reference error tags. Furthermore, it can diagnose the output of a system directly as well as identify remaining error types.
As shown in Table 1, ARETA encompasses several different errors based on extended ALC error types. ARETA's annotation process uses a combination of linguistic rules, morphological analysis, and alignment techniques to handle Arabic's intricacies, including its rich morphology and ambiguous orthography.

\begin{table}[t]
\caption{Error Type Categories}
\label{tab:error-categories}
\centering
\scalebox{0.9}{
\begin{tabular}{|l|l|l|}
\hline
\textbf{Category} & \textbf{Tag} & \textbf{Error}  \\ \hline
{Orthography} & OA & Alif, Ya \& Alif-Maqsura  \\ 
& OC & Char Order   \\
& OD & Additional Char   \\
& OG & Lengthening short vowels  \\
& OH & Hamza Error  \\
& OM & Missing char(s)  \\
& ON & Nun \& Tanwin Confusion  \\
& OR & Char Replacement  \\
& OS & Shortening long vowels \\
& OT & Ha/Ta/Ta-Marbuta Confusion   \\
& OW & Confusion in Alif Fariqa   \\ \hline
{Morphology} & MI & Word inflection  \\
& MT & Verb tense  \\ \hline
{Syntax} & XC & Case  \\
& XF & Definiteness   \\
& XG & Gender   \\
& XM & Missing word    \\
& XN & Number    \\
& XT & Unnecessary word   \\ \hline
{Semantics} & SF & Conjunction error  \\
& SW & Word selection error   \\ \hline
{Punctuation} & PC & Punctuation confusion  \\
& PM & Missing punctuation   \\
& PT & Unnecessary punctuation   \\ \hline
{Merge} & MG & Merge \\ \hline
{Split} & SP & Split  \\ \hline
\end{tabular}}
\end{table}

Aloyaynaa et al.~\cite{aloyaynaa2023arabic} analyzed Grammatical Error Detection (GED) in Arabic as a low-resource task. A7'ta~\cite{madi2019a7} was used along with publicly available datasets such as  QALB-14~\cite{mohit2014first}, QALB-15~\cite{rozovskaya2015second} and SCUT~\cite{solyman2022automatic}. Several approaches were employed for GED including token-level classification and sentence-level classification. A token-level classification was performed on individual tokens (words). Furthermore, sentence-level classification is used to determine whether a complete sentence is correct or incorrect. They used pre-trained transformer architectures, specifically mBERT and AraBERT, for GED classification. For the AraBERT and mBERT models, F1 scores of 85\% and 85\% respectively showed high performance metrics: Token-Level Classifications: 87\% for the AraBERT model and 86\% for the mBERT model. In terms of sentence-level classification, the AraBERT model scores 98\% and is 98\% accurate; in terms of the mBERT model, it scores 96\% and is 96\% accurate. Models built on monolingual data performed better than models trained on multilingual data for GED tasks in Arabic, demonstrating the effectiveness of using AraBERT pre-trained models.

While ARETA can be considered a multi-label classification, it requires both correct and incorrect text to align raw input sequences with reference sequences and identify error types. When both correct and incorrect text are present, ARETA is helpful, but it is not a predictive tool when only correct text is present. On another hand, As described in ~\cite{aloyaynaa2023arabic}, the main classification tasks involve binary classification for sentence-level classification, which categorizes sentences as either correct or incorrect. In contrast, token-level classification can be considered a multi-token classification task, since individual tokens (words) can be classified into correct and incorrect categories. While categorizing sentences as correct or incorrect aligns more closely with binary classification, the overall focus is on identifying which sentences are correct or incorrect. 
 \subsection{BERT Pre-trained Models}
Arabic pre-trained models have been used in various research projects to achieve outstanding results on a wide range of NLP tasks. The Arabic language is supported by some, while others support multiple languages, including Arabic. Table 2 presents an overview of the Arabic and multilingual BERT language model.

AraBERT~\cite{antoun2020arabert} was an early attempt to train a monolingual BERT model with Arabic news extracted from a variety of Arabic news sources. There are approximately 23 gigabytes of text in AraBERTv1 of the model, consisting of 77 million sentences and 2.7 billion tokens.  The newest version (AraBERTv2) is pre-trained with 77 gigabytes of text, 3.5 times more than the previous version. AraBERT consists of 12 transformer blocks, each containing 768 hidden units. As well as 110 million trainable parameters, it has 12 self-attention heads. An additional 12,000 sentences written in different dialects of Arabic were pre-trained on the model to support dialectical Arabic. SalamBERT was the name of this customized version of AraBERT\cite{husain2021leveraging}.

ARBERT~\cite{abdul2020arbert} is a masked language model that is pre-trained and aimed at Modern Standard Arabic (MSA). Several sources were used in the training model, including Wikipedia, news sources, and books. There are 6.2 billion tokens in it and it's about 61 gigabytes in size. ARBERT's structure comprises 12 layers of transformer blocks, 768 hidden units, and 12 self-attention heads. However, ARBERT has 163 million parameters that can be trained.

MARBERT~\cite{abdul2020arbert} was pre-trained on a massive amount of Twitter data that included both MSA and various Arabic dialects. It was also developed specifically for the Arabic language. Almost 128 gigabytes of text comprise the pretraining corpus, which contains 1 billion tweets. The number of tokens in the corpora is approximately 15.6 billion, which is more than double that of Version 2 of AraBERT. The MARBERT model is based on the multilingual BERT architecture, but it does not include the next-sentence prediction feature. The next-sentence prediction was omitted because tweets are too short, according to the model's developers. A total of 160 million parameters can be trained in MARBERT.

CAMeLBERT~\cite{inoue2021interplay} developed three pre-trained language models for Arabic: Modern Standard Arabic (MSA), dialectal Arabic, and classical Arabic, as well as a fourth model that was pre-trained on a mix of these three. The pretraining corpus of CAMeLBERT-MSA comprises almost 107 gigabytes. There are approximately 12.6 billion tokens in the corpora. The size of CAMeLBERT-DA is about 54 gigabytes and it contains 5.8 billion tokens. CAMeLBERT-CA contains 864 million tokens and is about 6 gigabytes in size. CAMeLBERT-Mix contains 17.3 billion tokens and is about 167 gigabytes in size. The CAMeLBERT model is based on the multilingual BERT architecture. A total of five NLP tasks were analyzed across 12 datasets, including sentiment analysis, part-of-speech tagging, dialect identification, named entity recognition (NER), and poetry classification. According to the average results, AraBERTv02 performed best across several subtasks. CAMeLBERT-Mix outperformed best models in dialectal contexts, but not consistently in other cases.

The DeBERTa model~\cite{he2020deberta} is an enhancement to the BERT and RoBERTa architectures, incorporating two innovation techniques: enhanced mask decoding and disentangled attention.  The enhanced mask decoder predicts masked tokens using absolute positions during pretraining. Disentangled attention allows for more effective attention weight calculations by representing each word with two vectors-one for content and another for position. Furthermore, the model's generalization on downstream tasks is improved using a new virtual adversarial training method. The DeBERTa model is mainly trained on English text, with 1.5 billion parameters, and has passed the human baseline of 89.816 on the SuperGLUE benchmark, exceeding human performance for the first time.  During the pre-training phase, DeBERTa uses about 78GB of data, which has been duplicated. This dataset contains several sources, including Wikipedia: 12GB, BookCorpus: 6GB, OPENWEBTEXT: 38GB (public Reddit content), and STORIES: 31GB (a subset of CommonCrawl).

In contrast to the traditional masked language modeling (MLM) approach, DeBERTaV3~\cite{he2021debertav3} uses a more efficient training objective called Replaced Token Detection (RTD) instead of the original DeBERTa. Gradient-disentangled embedding sharing is incorporated into this model to improve training efficiency. There are several variants of DeBERTaV3: DeBERTaV3large, DeBERTaV3base, DeBERTaV3small, and DeBERTaV3xsmall. Models developed using these methods have demonstrated superior performance on various NLU benchmarks, including GLUE, MNLI, and SQuAD v2.0. A multilingual version of DeBERTaV3, mDeBERTaV3base, has also been developed to outperform other multilingual models like XLM-Rbase across a large dataset (CC100). This makes DeBERTaV3 a versatile model suitable for various languages and tasks in natural language processing.

BERT has not been applied to the classification of Arabic text in grammatical error correction tasks, so studies are needed to determine how to classify multi-label Arabic text. To the best of our knowledge, there was no Arabic research or model for predicting types of grammatical errors included in correct Arabic texts at the time of this study. Based on these considerations, we developed a pre-trained model that can predict multi-label classifications for Arabic text grammatical errors.

\begin{table*}[t]
\centering
\setlength{\tabcolsep}{5pt} 
\renewcommand{\arraystretch}{1.5} 
\caption{An overview of the Arabic and multilingual BERT pre-train language models.}
\label{tab:models}
\resizebox{\textwidth}{!}{ 
\begin{tabular}{|m{2.5cm}|m{2cm}|m{2cm}|m{2.2cm}|m{1.5cm}|m{1.5cm}|m{1.5cm}|m{1.5cm}|m{5cm}|m{3cm}|}
\hline
\textbf{Model} & \textbf{Architecture} & \textbf{Parameters} & \textbf{Language} & \textbf{Size} & \textbf{\#Word} & \textbf{Token} & \textbf{Vocab} & \textbf{Pre-train Data} & \textbf{Tasks} \\ 
\hline
AraBERTv01 & BERT-base & 110 M & Arabic (MSA) & 24GB & - & SP & 60K & (1) The 1.5 billion words Arabic Corpus~\cite{el20161}. (2) OSIAN Arabic News Corpus~\cite{zeroual2019osian}. & Sentiment Analysis (SA), Named Entity Recognition (NER), Question Answering (QA) \\ 
\hline
AraBERTv02 & BERT-base & 135 M & Arabic (MSA) & 77GB & 8.6B & WP & 60K & (1) Arabic Corpus~\cite{el20161}. (2) OSIAN Corpus~\cite{zeroual2019osian}. (3) OSCAR~\cite{suarez2020monolingual}. (4) Wikipedia \footnote{\url{https://archive.org/details/arwiki-20190201}}. (5) Assafir news. & - \\ 
\hline
CAMeLBERT-MSA & BERT-base & 110 M & Arabic (MSA) & 107GB & 12.6B & WP & 30K & Arabic Gigaword \footnote{\url{https://catalog.ldc.upenn.edu/LDC2011T11}}, El-Khair Corpus~\cite{el20161}, OSIAN Corpus~\cite{zeroual2019osian}, Arabic Wikipedia \footnote{\url{https://archive.org/details/arwiki-20190201}}. & NER, POS tagging, Sentiment Analysis, Dialect Identification, Poetry Classification \\ 
\hline
CAMeLBERT-DA & BERT-base & 110 M & Arabic (DA) & 54GB & 5.8B & WP & 30K & Range of dialectal corpora~\cite{inoue2021interplay}. & - \\ 
\hline
CAMeLBERT-CA & BERT-base & 110 M & Arabic (CA) & 6GB & 847M & WP & 30K & OpenITI Corpus (v1.2) \footnote{\url{https://github.com/OpenITI/RELEASE}}. & - \\ 
\hline
CAMeLBERT-Mix & BERT-base & 110 M & Arabic (MSA/DA/CA) & 167GB & 17.3B & WP & 30K & Combination of previous CAMeLBERT data. & - \\ 
\hline
ARBERT & BERT-base & 163 M & Arabic (MSA) & 61GB & 5.6B & WP & 100K & (1) 1,800 Arabic books (Hindawi) \footnote{\url{https://www.hindawi.org/books/}}. (2) El-Khair~\cite{el20161}. (3) Gigaword \footnote{\url{https://catalog.ldc.upenn.edu/LDC2011T11}}. (4) OSCAR~\cite{suarez2019asynchronous}. (5) OSIAN~\cite{zeroual2019osian}. (6) Wikipedia Arabic \footnote{\url{https://github.com/attardi/wikiextractor}}. & SA, SM, TC, DI, NER, QA \\ 
\hline
MARBERT & BERT-base & 163 M & Arabic (MSA + DA) & 128GB & 15.6B & WP & 100K & 1B Arabic tweets from an in-house dataset~\cite{abdul2020arbert}. & - \\ 
\hline
DeBERTa & DeBERTa-base & 1.5 B & Multilanguage & 78GB & - & WP & 128K & Wikipedia \footnote{\url{https://dumps.wikimedia.org/enwiki/}}, BookCorpus~\cite{zhu2015aligning}, OPENWEBTEXT~\cite{gokaslanopenwebtext}, STORIES~\cite{trinh2018simple}. & Acceptability, SA, NLI, Paraphrase Detection, QA \\ 
\hline
\end{tabular}
}
\end{table*}

\subsection{Text-to-Text Transformer}

A state-of-the-art model for natural language processing (NLP) tasks, Text-to-Text Transfer Transformer (T5)~\cite{raffel2020exploring} can be applied to a variety of NLP tasks by treating them as text-to-text tasks. In addition to classifying, summarizing, and translating, the model uses a unified architecture and loss function. T5 utilizes a transformer architecture that uses encoder-decoder structures to produce coherent and fluent results. This architecture is composed of two parts: The encoder generates a contextual representation of the input sequence. In producing each entry of the output, the self-attention mechanism uses a fully visible attention mask to attend to any entry of the input. The decoder generates the output sequence from the encoder's output. This technique uses both causal masking and encoder-decoder attention to maintain autoregressive properties while at the same time attending to the encoder's output. Multiple benchmarks have been run on the model, and it performs better when scaled to various sizes, with the largest variant, T5-11B, containing 11 billion parameters. A powerful dataset known as the Colossal Clean Crawled Corpus (C4) was used to pre-train the model, making it able to learn from a wide range of texts. Although T5 performed well on several tasks, including CNN/Daily Mail summarization benchmarks, it performed poorly on some translation tasks, perhaps due to its reliance on an English-only dataset.

mT5~\cite{xue2020mt5} is an enhancement of T5 that utilizes GeGLU nonlinearities and scales both feed-forward dimensions and model dimensions in larger models. The encoder-decoder model supports generative tasks such as abstractive summarization and dialog, which is different from encoder-only models such as XLM-R. As a result of training more than 250,000 words, the mT5 model has several sizes, such as Small (300M parameters), Base (580M), Large (1.2B), XL (3.7B), and XXL (13B). As part of the pre-training of the mT5 model, a large dataset named mC4, which contains over 100 different languages of text, is used. In the training process, data is sampled from each language to balance the representation of languages with high resources and languages with low resources.  Several classification and question-answering tasks have been performed using the mT5 model, and it has proven to be state-of-the-art. Various training strategies are employed, including in-language multitask training ( utilizing gold data within the target language), translate-train ( utilizing machine translation from English ), and zero-shot transfer ( utilizing only English data for fine-tuning). Cross-lingual representation learning is influenced by model capacity, with larger models outperforming smaller ones, particularly in zero-shot scenarios.

In ByT5~\cite{xue2022byt5}, vocabulary does not need to be built or tokenized like in mT5 and the NLP pipeline is simplified. The ByT5 architecture supports byte-level processing, so vocabulary parameters are reallocated to the transformer layers, improving model efficiency. ByT5 models come in different sizes (300M, 582M, 1.23B, 3.74B, and 12.9B), all with varying hidden sizes and feed-forward dimensions. A multilingual task can be effectively handled by both ByT5 and mT5. In multitasking settings where gold training data is available, ByT5 has shown competitive performance across a variety of languages. Multiple benchmarks demonstrate its ability to manage tasks in multiple languages, surpassing mT5. ByT5 models range in parameter count from 300 million in the Small version up to 12.9 billion in the XXL version. A comparison of mT5 models reveals that the Base model has 582 million parameters, while the Large model has 1.23 billion. The ByT5 design improves performance and reduces system complexity, making it a viable alternative to token-based models like mT5, especially for applications without significant latency issues.

AraT5~\cite{nagoudi2021arat5} utilizes the T5 (Text-to-Text Transfer Transformer) encoder-decoder architecture, in particular the T5Base encoder-decoder. An AraT5MSA variant was developed (trained on Modern Standard Arabic data), an AraT5TW variant (trained on Twitter data), and an AraT5 variant (trained on both MSA and Twitter data). It is estimated that there are approximately 220 million parameters in each model, which is composed of 12 layers, 12 attention heads, and 768 hidden units. Various datasets were used to pre-train the models, including unlabeled MSA and Twitter data. A self-supervised (denoising) objective was used to train the model, with 15\% of tokens randomly masked to reassemble the original sequence. Besides the code-switched datasets, the data also includes monolingual French and English translations of Algerian and Jordanian Twitter. An Arabic language GENeration (ARGEN) benchmark was used to evaluate the models, which included seven tasks: machine translation, code-switched translation, text summarization, news headline generation, question generation, paraphrasing, and transliteration. Over 52 out of 59 test sets, the models outperformed the mT5 model with an 88.14\% performance rate. On the Arabic language understanding benchmark ARLUE, they also set new state-of-the-art (SOTA) results.
\section{Data}
This section is to provide an overview of the parallel data that will primarily be used to train our methodologies models and monolingual corpus used to generate large synthetic data.
\begin{table}[t]
\centering
\setlength{\tabcolsep}{5pt} 
\renewcommand{\arraystretch}{1.5} 
\caption{Arabic corpus used in ArabGEC}

\begin{tabular}{|c||c||c||c||c||c|}
\hline
Corpus & Split  & Sentences & Words & Level & Domain  \\ \hline
QALB-14 &
 \begin{tabular}[c]{@{}c@{}}Train\\ Dev\\ Test\end{tabular} &
 \begin{tabular}[c]{@{}c@{}}19.4K\\ 1K\\ 948\end{tabular} &
 \begin{tabular}[c]{@{}c@{}}1M\\ 54K\\ 51K\end{tabular} &
 \begin{tabular}[c]{@{}c@{}}L1\\ L1\\ L1\end{tabular} &
 \begin{tabular}[c]{@{}c@{}}Comments\\ Comments\\ Comments\end{tabular} \\ \hline
QALB-15 &
 \begin{tabular}[c]{@{}c@{}}Train\\ Dev\\ Test-L1\\ Test-L2\end{tabular} &
 \begin{tabular}[c]{@{}c@{}}310\\ 154\\ 158\\ 940\end{tabular} &
 \begin{tabular}[c]{@{}c@{}}43.3K\\ 24.7K\\ 22.8K\\ 48.5K\end{tabular} &
 \begin{tabular}[c]{@{}c@{}}L2\\ L2\\ L2\\ L1\end{tabular} &
 \begin{tabular}[c]{@{}c@{}}Essays\\ Essays\\ Essays\\ Comments\end{tabular} \\ \hline
ZAEBUC & No spilt & 214 & 33,376 & L1  & Essays  \\ \hline

\end{tabular}%

\end{table}
\subsection{Parallel Corpus}

The purpose of this section is to provide information about the data that will primarily be used to train our methodologies models. Table 3 shows how data are split, number of sentences, words, levels, and domains.

\textbf{QALB-14 and QALB-15 }. The QALB-14~\cite{mohit2014first} corpus contains Al Jazeera News comments written in Modern Standard Arabic by native speakers. Arabic speakers of all backgrounds were addressed in QALB-2015, including native Arabic speakers and non-native Arabic speakers. QALB-15~\cite{rozovskaya2015second}, which collects texts from the Arabic Learner Corpus (ALC)~\cite{alfaifi2012arabic} and the Arabic Learner Written Corpus (ALWC)~\cite{farwaneh2012arabic} of Arabic learners, includes texts extracted from the two learner corpora. There are three phases to the annotation process: automatic preprocessing, automatic spelling corrections, and human annotations. The spelling correction was automated using the morphological analysis~\cite{habash2005arabic} and disambiguation system MADA (version 3.2)~\cite{habash2009mada+}. Besides spelling and punctuation, annotators had to fix morphology, syntax, and dialect errors. In QALB-14, there are 21,396 sentences, and in QALB-15, there are 1,533 sentences. 

\textbf{ZAEBUC} ~\cite{habash2022zaebuc} contains annotated texts in Arabic and English by first-year Zayed University students. It contains short essay bilingual corpora that are matched to two writers, one who writes in their native language and one who writes in their second language. Four steps were involved in the creation of the corpus. First, ZU's IRB board had to approve the study, followed by contacting faculty teaching the targeted courses. All participating students provided written consent. Additionally, manual text corrections and CEFR annotations were completed independently. Depending on the results, morphological annotation was followed by text correction. Manually correcting the semi-automated annotations was the last step. The corpus consisted of 214 sentences in total, which was relatively small.
\subsection{Monolingual Corpus}
The following paragraphs describe the corpus used in the synthetic data generation process.
\textbf{1.5 billion words} corpus~\cite{el20161}. The corpus contains over 1.5 billion words from ten major Arabic news sources, including newspapers and news agencies in eight Arab countries. The sources are as follows: Alittihad (United Arab Emirates), Echorouk Online (Algeria), Alriyadh (Saudi Arabia), Almustaqbal (Lebanon), Almasryalyoum (Egypt), youm7 (Egypt), Saba News Agency (Yemen). The data collection period for these sources ranged from January 2002 to June 2014, depending on the source. Articles published online during these periods are included in the corpus. There are many topics covered in the corpus, including politics, literature, the arts, technology, sports, and the economy. As a result of this diversity, it is an excellent resource for researchers in the fields of natural language processing and information retrieval. Furthermore, the corpus contains more than five million articles and 1.5 billion words, with more than three million unique words. Two different encoding schemes were used for the corpus: Windows CP-1256 and UTF-8.  The corpus also contains two markup languages: SGML (Standard Generalized Markup Language) and XML (Extensible Markup Language). 

\textbf{Open Source International Arabic News (OSIAN) corpus }\cite{zeroual2019osian}. An annotation-based corpus of Arabic news articles aimed at addressing the limited resources nature of Arabic in linguistic computation. The corpus includes 3.5 million articles, 37 million sentences, and approximately 1 billion tokens annotated with metadata and linguistic information from international Arabic news websites. The data is encoded in XML format, which facilitates structured storage. The corpus includes metadata about each article, including its source domain name, URL, and extraction date. For natural language processing, the articles also provide linguistic information about each word, including its lemma and part of speech tags. In addition to covering a wide range of topics and sources, the corpus is designed to be a balanced representation of international Arabic news. A separate directory for each country contains the articles, which are lemmatized and tagged for correct part-of-speech in XML files. Furthermore, statistics on word length distribution and word frequency lists are included in the corpus, which can help analyze the linguistic characteristics of text. 

\section{Methodology}
As shown in Figure 1, we use two models to generate our data: an error tagging and a synthetic data generation model. In the error tagging model, we used the deBERTav3 to assign and predicate an error label that may appear in incorrect sentences based on a subset of the 26 Arabic Error Type Annotation tool (ARETA) error tags. Then, a synthetic data generation model used an AraT5 model to generate an incorrect version of the data as a result of concatenating this label with the correct sentence.
\begin{figure*}[t]
    \centering
    \includegraphics[width=0.9\linewidth]{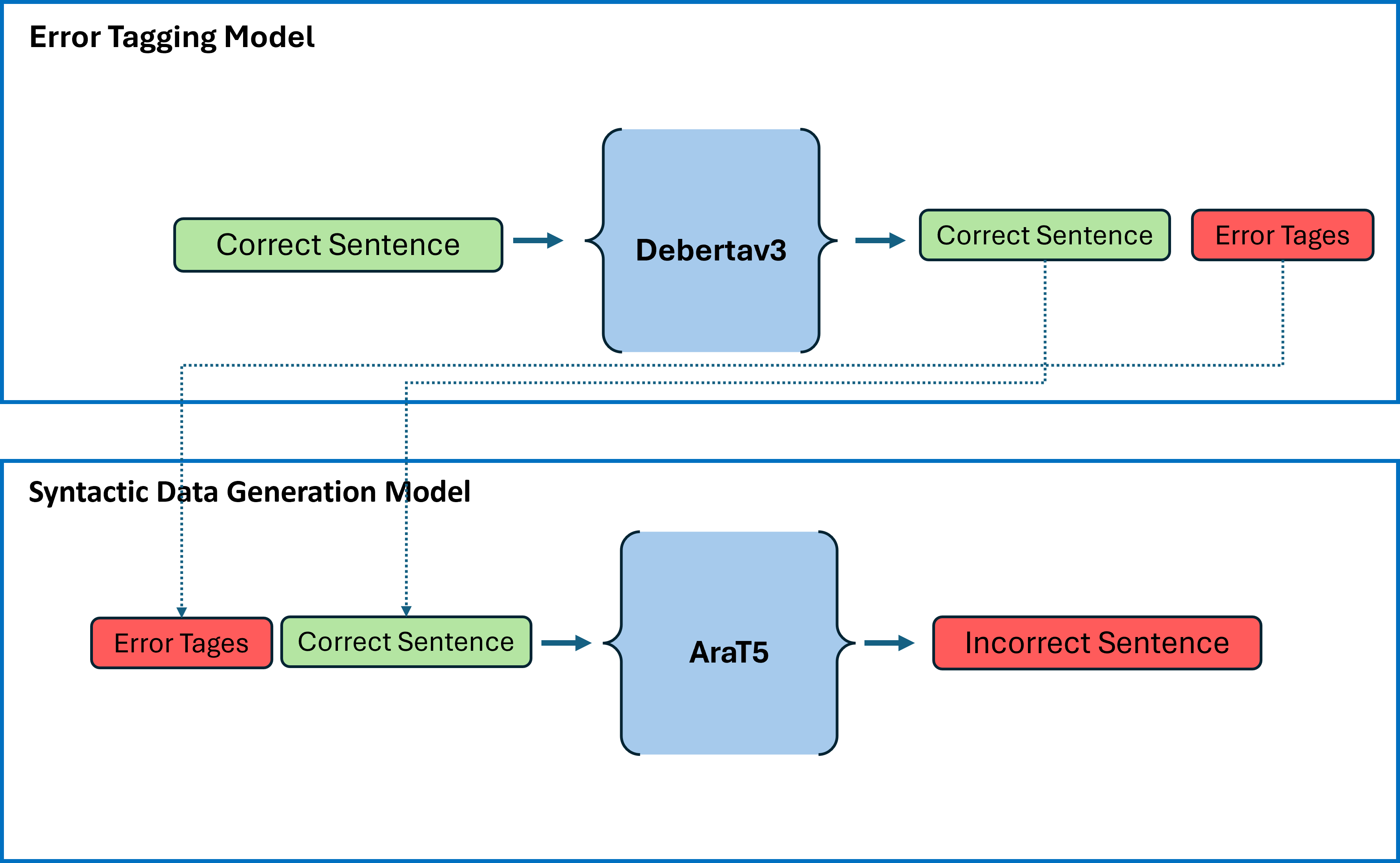}
    \caption{The error tagging model and the synthetic data generation model.}
    
\end{figure*}

\subsection{Error Tagging Model}
The error-tag model is a multi-label classification task that predicts the types of errors that may occur in a correct sentence. Our motivation for developing such a model was the availability of many correct and error-free data sources in news, newspapers, magazines, and books, as well as many monolingual corpora exist, but there are lack of parallel data sources for training grammatical error correction (GEC) tasks. In our cases, Arabic is considered a low-resource language and has one parallel corpus (QALB-14 and QALB-15) for GEC tasks. In addition, the Arabic language lacks a model for predicting error type as a multi-label classification task. Based on what we described in the related work section, most current research needs both correct and incorrect data to determine the error type or classify it as either correct or incorrect sentence or token, so this model will help identify errors in correct sentences and allow for further parallel data generation.

Error tagging models were trained on QALB-14 QALB-15 shard data and ZAEBUC, the only available grammatical error data. Our goal is to train our model on real data so that it can predict human-like erroneous sentences. To train, we used the QALB-14 train set, the QALB-15 train set, and ZAEBUC, which totaled 19935 samples. QALB-14 development set and QALB-15 development set were used for validation, totaling 1171 samples. Finally, we tested using the QALB-14 test set, QALB-15-L1, and QALB-15-L2 test sets, which contained 2,046 samples in total. Then, we used the ARETA tool to identify error tags for each sentence. According to Table 1, ARETA contains 26 error types in 7 classes. While ARETA can identify some error types effectively and efficiently, it will not identify some other error types, so they will be marked with Xs or UNKs. Hence, we manually identified Xs or UNKs error types to appropriate tags.

To generate multi-label classification for Arabic text. We train and compare five MSA pre-trained BERT models AraBERTv2~\cite{antoun2020arabert}, CAMelBERT-MSA~\cite{inoue2021interplay}, ARBERTv02~\cite{abdul2020arbert}, MARBERTv02~\cite{abdul2020arbert} and Debertav3~\cite{he2021debertav3} from Hugging-Face~\footnote{\url{https://huggingface.co/}} . We fine-tuned them using a multi-label classification objective. Therefore, multiple error tags can be predicted for the same input sentence at the same time. To accomplish this, each of the output layer nodes can be activated with the sigmoid activation function, along with binary cross-entropy loss.

Each model was trained independently on the same data and the results were reported in section 7. Among the models, Debertav3 provides a high result in terms of f1-score and was able to infer all types of errors. Following ARABERTv02, then CAMELBERT-MSA, then ARBERTv02, and lastly MARBERTv02. We therefore mostly rely on the DeBERTa pre-train model to predict error types for each sentence in the error tagging model.

Since we are dealing with a multi-label classification problem, one sentence can often be tagged with more than one error, as the labels in (QALB-14, QALB-15, and ZAEBUC) are highly imbalanced as illustrated in Figure 2. The model may not predict labels that occur infrequently if there is no correction, resulting in the model predicting labels that occur frequently instead.
\begin{figure*}[t]
    \centering
    \includegraphics[width=1.1\textwidth]{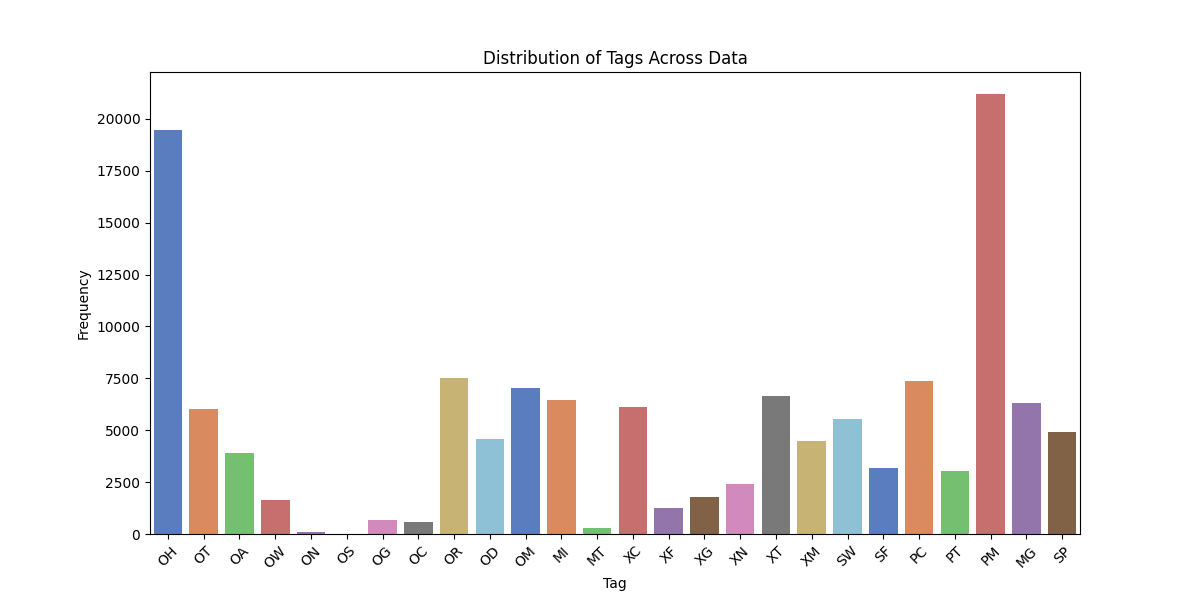}
    \caption{Label imbalances in (QALB-14, QALB-15, and ZAEBUC). Over half of the data contain some labels, while almost none occur, resulting in large label imbalances.}
    
\end{figure*}

To address imbalances in class distribution, distribution-balanced loss (DB)~\cite{huang2021balancing} is used for multi-label text classification. A distribution-balanced loss aims to address this issue by balancing the contributions from all samples. Each sample's loss is re-weighted according to its frequency of occurrence in the dataset, which is determined by the inverse frequency of occurrence of that sample. A lower weight is explicitly assigned to instances of negative data that are easy to classify by the DB loss function. In multi-label classification, some labels may be easier to predict than others, causing poor performance on more challenging labels if too much emphasis is placed on these easier labels. As a result, less frequent classes contribute more, and more frequent classes contribute less, thus balancing the overall distribution of classes. Models with long-tailed class distributions have been shown to perform better using this technique. To produce more balanced predictions, the model reduces the impact of frequent classes and increases the impact of infrequent classes.

As part of our effort to improve our model, we tuned the threshold during training, validation, and testing. We calculated the dependencies of metrics such as F1-score, precision, and recall on the threshold level, and we selected the threshold based on the highest metric score. The objective is to find a threshold where the probability theory for two values (0 or 1) is at its maximum performance based on the F1 Score measure. Several thresholds are tried within a range close to the default value (0.5), where the supervision of each sticker and part is calculated using F1 Score, Precision, and Recall. A threshold is determined by determining the F1 score that is highest for each strip. A balance between precision and recall is achieved by returning this optimal threshold to improve the outstanding model for classifying classes.

\subsection{Synthetic Data Generation Model}
To generate synthetic data, we mainly used a back translation model. A reversed GEC model is trained to create ungrammatical sentences from grammatical ones. Hence,
we fine-tuned AraT5 two times to generate erroneous sentences. In the first fine-tuning process, the model input is the incorrect sentence and the target is the correct sentence. We are using a pre-trained version of "AraT5v2-base-1024" from the Happy Transformer package. As training data, we used QALB-14, QALB-15, and ZEABAQ, and as development data, QALB-14, QALB-15. 

Our model is then further fine-tuned for a second time to generate syntactic data. We used the same data as in GEC but reversed the training process. Our input strings are prefixed with grammar\_error as task conditioning for AraT5. There are 26 types of errors represented by bracketed 26 characters consisting of "a" and "b". As described in our previous subsection, these tags are generated by the error tagging model. In the parallel corrupted sentence, "b" indicates the presence of error type i. In contrast, "a" is the opposite. Figure 3 shows how we can generate and represent sentences with multiple errors as a result. In the final step, we append our corrupted sentence to these tags. As a result of the model, we get the corrupted sentence as an output. The batch size is 8, the learning rate is 2e-5, and the number of epochs is 300.

\begin{figure*}[t]
    \centering
    \includegraphics[width=.9\textwidth]{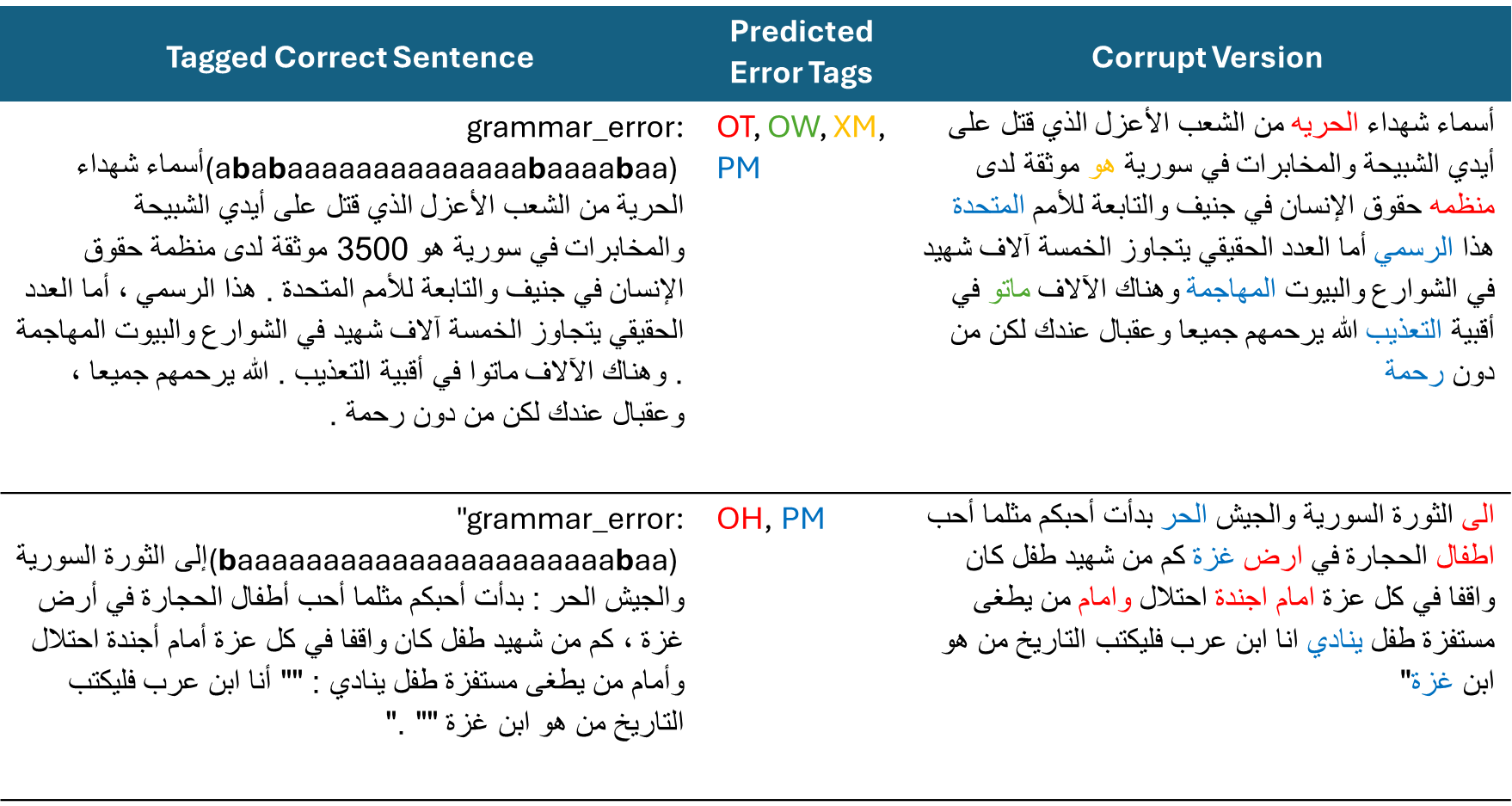}
    \caption{Examples of data generated by our AraT5 synthetic data generation model. We use our error tagging model to predict error tags. An error that is highlighted or bolded in a corrupt sentence is caused by the corruption model and is highlighted in that color.}
    
\end{figure*}

\subsection{Data Generation}
By using the 1.5 billion word and OSAIN datasets, we generate 30,219,310 synthetic sentence pairs after training our error tagging model and syntactic data generation models. To generate parallel corpora, it must first be a preprocessed monolingual corpus, as described in section 4.2. A 1.5 billion word corpus, consisting of XML files, was first analyzed by extracting only text tags and storing them in text files for easier processing. To further process some files, we split them into two or more files as needed due to their large size. Conversely, the OSIAN corpus text already exists in separate text files. However, it contains a sequence number at the beginning of each sentence, so we removed it. Then, we discovered that both corpora contained spelling and grammatical errors. Several spelling errors were corrected manually by replacing fatha Tanween with the letter preceding the alif, replacing the Ya with an Alif-Maqsura and vice versa, and modifying hazmat errors. The punctuation was also corrected by removing duplicate punctuation, adjusting the shape of punctuation like commas and adjusting the punctuation between numbers to ensure that the text was clear and concise. Grammatical errors were not corrected because they needed to be annotated by humans. We then pre-processed the data using natural language processing techniques, including removing empty lines, sentences with less than 10 words, non-UTF8 encoding, and over-space.  

In our final step, we will use the syntactic data generation models we described in the 5.2 subsections to generate our corpus using the prefix generated in the 5.1 subsections and shown in Figure 3.

\subsection{Grammatical Error Correction}
Our synthetically generated data was used to train a GEC model using the AraT5 model to evaluate the effectiveness of performance gain. For evaluating our model with other Arabic synthetic data generating models, we used synthetic data from~\cite{solyman2023optimizing}, which is the most recent Arabic study that generated synthetic data. Data is organized into seven folders, each of which contains a specific type of error. The errors were generated by Misspelling (producing spelling errors), Swap (changing the order of target words), Reverse (transposing left-to-right words into right-to-left words), and Replace (aligning target and source words), which achieved the best performance, respectively. Token (adding UNK tokens to target sentences), and Source (replacing the source sentence with the target sentence) have achieved lower performance. There is a file for training and another for validation in each folder. The best-performing data is in the eighth file, which contains (spelling + swap + reverse). Data was collected in two files, one for training and one for validation. The number of training data was 333908 and the number of validation data was 18206. These synthetic data were generated from QALB-14 and QALB-15, which are derived from the QALB-14 and QALB-14 data. To generate our synthetic data, we extract only the correct text from~\cite{solyman2023optimizing} synthetic data, then predict the expected error tag using an error tagging model, and then produce the incorrect text by using our synthetic data generation. Our evaluations are based on using the QALB-14-L1, the QALB-15-L1, and the QALB-15-L2 test sets.

Two models are fine-tuned, one that uses~\cite{solyman2023optimizing} data, and the other that uses the same correct sentences but we generate incorrect sentences by using our models. We used batch size 8 for AraT5. We set the learning rate to 5e-5. After 23 train epochs, we fine-tuned by using the QALB-14-L1 and QALB-15-L1 training data for two epochs.

\section{Experiments}
In this section, we describe our evaluation metrics and experimental setup used to evaluate our error tagging model and synthetic data generation model.
\subsection{Evaluation Measures}
Our tagging model is considered a multi-label classification task. A multi-label classification allows multiple labels to be assigned simultaneously to an output, whereas a single-label classification allows only one label to be assigned. Thus, evaluating the performance of multi-label classification is more challenging and complicated~\cite{wu2017unified} and requires a different approach than evaluating single-label classification performance; multiple metrics are recommended. Several multi-label classification evaluation measures are described and categorized in~\cite{wu2017unified}. This work was evaluated using Hamming loss, micro-precision, micro-recall, micro-F1, and micro-F0.5.

The precision of a classification prediction is the percentage of the correct positive classification predictions. A recall score is defined as how many positive classes were correctly predicted, while an F1-score, which is the recall and precision harmonic mean, is an evaluation measure most commonly used. F0.5 is calculated by averaging precision and recall (given a threshold value). The F0.5 score places more emphasis on precision than recall, unlike the F1 score, which gives equal weight to both. False Positives should be weighted more heavily than False Negatives in cases with higher precision. For precision, recall, F1-score, and F0.5 score, we apply the following definitions to several true positives TP, true negatives TN, false positives FP, and false negatives FN:

\begin{equation} 
Precision = \frac{TP}{TP+FP}
\end{equation} 
\begin{equation} 
Recall = \frac{TP}{TP+FN}
\end{equation} 
\begin{equation} 
F1-Score = \frac{2 \times TP}{2TP+FP+FN}
\end{equation} 
\begin{equation} 
F0.5-Score = 1.25\times \frac{(Precision)(Recall)}{0.25\times(Precision+Recall)}
\end{equation} 

Multiple labels can be evaluated via micro-averaging, macro-averaging, and Weighted-averaging~\cite{yang1999evaluation}. Metrics for macro-averaging are calculated independently for each label before averaging them all. By contrast, micro-averaging aggregates the number of hits and misses before calculating the desired metric once~\cite{herrera2016multilabel}. Therefore, the final measure calculation takes into account the disparate data distribution associated with each label as well as the weight assigned to each label. In a weighted average, all per-class scores are averaged, adjusting for the support that each class receives. Although micro-averaging is commonly used for evaluating performance when the data is unbalanced~\cite{yang2009effective}, the following are the definitions of micro-F1, micro-F0.5, micro-precision, and micro-recall:
\begin{equation}
\text{Micro-Precision} = \frac{\sum_{i=1}^{n} TP_i}{\sum_{i=1}^{n} (TP_i + FP_i)}
\end{equation}
\begin{equation}
\text{Micro-Recall} = \frac{\sum_{i=1}^{n} TP_i}{\sum_{i=1}^{n} (TP_i + FN_i)}
\end{equation}
\begin{equation}
\text{Micro-F1} = 2 \cdot \frac{\text{Micro-Precision} \cdot \text{Micro-Recall}}{\text{Micro-Precision} + \text{Micro-Recall}}
\end{equation}
\begin{equation}
\resizebox{.6\hsize}{!}{$\text{Micro-F0.5} = 
 (1 + 0.5^2) \cdot \frac{\text{Micro-Precision} \cdot \text{Micro-Recall}}{(0.5^2 \cdot \text{Micro-Precision}) + \text{Micro-Recall}}$}
\end{equation}
The Hamming loss is a measure of the percentage of incorrect predictions made across an entire set of labels. A model's performance is higher when its hamming loss is lower. Hamming loss can be calculated as follows:
\begin{equation}
\text{Hamming Loss} = \frac{\text{FP} + \text{FN}}{\text{FP} + \text{FN} + \text{TP} + \text{TN}}
\end{equation}

Where:
- \(\text{FP}\) is the number of false positives,
- \(\text{FN}\) is the number of false negatives,
- \(\text{TP}\) is the number of true positives,
- \(\text{TN}\) is the number of true negatives.

To evaluate our synthetic data generation model and grammatical error correction model, we used MaxMatch~\cite{dahlmeier2012better} to evaluate the performance in a shared task by measuring word-level edits compared to golden target sentences and to report precision, recall, F1 and F0.5 scores during training using different scenarios. In addition to providing insights into a GEC system's strengths and weaknesses, these metrics evaluate different aspects of the system's performance.

Furthermore, BLEU-4 scores were applied to evaluate machine correction, and a case study was provided to evaluate human correction. A BLEU-4 score indicates how similar the corrected output is to the reference output. Up to a maximum n-gram size of 4, it's calculated by comparing the corrected output to the reference output:

\begin{equation}
\text{BLEU-4} = \text{BP} \times \exp \left( \sum_{n=1}^{4} w_n \cdot \log P_n \right)
\end{equation}

Where:

\begin{equation}
\text{BP} =
\begin{cases}
1, & \text{if } c \geq r \\
\exp(1 - \frac{r}{c}), & \text{if } c < r
\end{cases}
\end{equation}

\begin{itemize}
    \item \( c \) represents the candidate translation's length.
    \item \( r \) represents the length of the reference translation.
    \item \( w_n \) indicates the weight of each n-gram (usually \( w_n = \frac{1}{4} \) for BLEU-4).
    \item \( P_n \) measures the precision of the n-grams by comparing matching n-grams to all n-grams of the candidate.
\end{itemize}

\subsection{Experiment Setup}
In an error tagging model, we use the AraBERTv02, the CAMELBERT-MSA, the ARBERTv2, the MARBERTv2, and the DeBERTa-v3-based architectures. The aim is to identify multiple tags (or labels) associated with a text sequence input. We implemented the Hugging Face Transformers library's base version in our experiments. The Hugging Face Transformers library is a well-known, publicly accessible library~\cite{wolf2020transformers}. All of the pre-trained model weights have been included in the Hugging Face Transformers library. With such libraries, pre-trained models can be unified, and pretraining can be performed with as little computational resource usage as possible. Since we only have one pre-trained model, DeBERTa-v3, 'Microsoft/debertav3-base', for multilingual DeBERTa models in Arabic, we use it in the experiment. As a comparison, we employ Arabic MSA AraBERTv02, CAMELBERT-MSA, ARBERTv2, and the MARBERTv02 models named ''aubmindlab/bert-base-arabertv02", "CAMeL-Lab/bert-base-arabic-camelbert-msa", "UBC-NLP/ARBERTv2" and ''UBC-NLP/MARBERTv2''.

To maximize training efficiency, several hyperparameters are set using the TrainingArguments API. In both training and evaluation, batch size is set to 8 samples per GPU. The learning rate is 2e-5, with a weight decay of 0.1. There is a checkpoint saved every 1000 steps, and the learning rate is evaluated every 1000 steps. A total of 120 epochs is required to ensure convergence. Training on GPUs can be sped up using mixed precision (fp16). Early stopping is incorporated with a patience of 20 steps and an improvement threshold of 0.001. Depending on the evaluation loss, up to three checkpoints are saved during the training process. Sigmoid activation and threshold are used in the model to generate predictions. 

For handling the class imbalance in the dataset, we use a custom loss function (ResampleLoss)~\cite{huang2021balancing}. With this loss function, class importance is dynamically adjusted through a focal loss component. Weights are assigned based on class frequencies and sample counts. The model optimizes performance across frequent and minority classes using binary cross entropy (BCE), focal loss, and class-balanced loss.  Focus loss is one of the most important components of this model. Focus loss underlines hard-to-classify data by modulating it with factors such as gamma (minimizing the influence of well-classified data) and alpha (minimizing the influence of positive labels). A class-balanced loss (CB) also ensures that minority classes contribute meaningfully to training by applying a correction factor based on class frequency. To stabilize predictions for imbalanced datasets, the function uses nonlinear functions, such as logit regularization, which further enhances optimization. 

PyTorch and Hugging face Transformers are used to implement the experiment within the Aziz Supercomputer Execution Environment, which leverages GPUs for efficient training. It also facilitates fine-tuning and testing with new data by ensuring the reproducibility of results and models. 
\section{Experimental Results and Discussion}
The results of the models, the error tagging model, and the model for generating synthetic data are provided in this section.
\subsection{Performance of Error Tagging Model}
Our error tagging model was evaluated on the QALB-14 and QALB-15 test set as shown in Table 4 and by combining the development data for QALB-14 and QALB-15, and then evaluating on the following measures: F1, F0.5, Precision, Recall, and Hamming losses as shown in table 4.

Table 4 indicates that of all models tested in the QALB-14 Dataset, DeBERTa-v3-base achieved the highest F1 (0.9406), F0.5 (0.94154), Precision (0.94216), Recall (0.93905), and Hamming Loss (0.02711), reflecting its high effectiveness and accuracy. The second-best performance, AraBERTv02 achieved an F1 of (0.91048) and a Hamming loss of (0.04001). CameLBERT-MSA is good, but a little less good than AraBERTv02, with F1 (0.90479) and Hamming loss (0.04263). Both ARBERTv02 and MARBERTv02 have lower performance with Hamming Loss (0.06422 and 0.0647) compared to previous models.

Additionally, DeBERTa-v3-base leads with the lowest Hamming loss (0.0218) in the QALB-15-L1 dataset indicating that it is performing well. In terms of the F1 and Hamming loss, AraBERTv02 ranked second with 0.91782 and 0.03423, respectively. In terms of F1 (0.90663) and Hamming loss (0.03867), CameLBERT-MSA performed well, but less well than AraBERTv02. The ARBERTv02 and MARBERTv02 both had relatively poor Hamming losses (0.0553 and 0.06048, respectively).

The DeBERTa-v3-base version continues to perform well in QALB-15-L2 data, exhibiting a high F1 of 0.81628 and a low Hamming loss of 0.18136. Second place is taken by ARABERTv02 but with a hamming loss of 0.18695. There was a significant difference in performance between CAMELBERT-MSA, ARBERTv02, and MARBERTv02, with Hamming losses that ranged from 0.19109 to 0.20497, indicating the difficulty of the data in this set.

Due to our model being trained in L1 data, QALB-15-L2 showed relatively lower performance than the other dataset. In the L1 dataset, errors made by native Arabic speakers may be fewer or simpler than errors made by learners. L2 datasets, on the other hand, usually contain more complex and frequent errors, as they are made by people still learning Arabic, making the necessary corrections more challenging.

\begin{table*}[t]
\centering
\setlength{\tabcolsep}{5pt} 
\renewcommand{\arraystretch}{1.5} 
\caption{A comparison of different BERT models based on evaluation metrics on separate QALB test sets.}

\scriptsize
\resizebox{\textwidth}{!}{%
\begin{tabular}{|c|c|c|c|c|c|c|}
\hline
\textbf{Data} & \textbf{Model} & \textbf{F1} & \textbf{F0.5} & \textbf{Precision} & \textbf{Recall} & \textbf{Hamming - loss} \\ \hline
   QALB-14 Test Set & AraBERTv02 & 0.91048 & 0.92317 & 0.93183 & 0.89009 & 0.04001 \\ \cline{2-7} 
 & CAMeLBERT-MSA & 0.90479 & 0.91639 & 0.92429 & 0.88609 & 0.04263 \\ \cline{2-7} 
 & ARBERTv02 & 0.85767 & 0.86451 & 0.86914 & 0.8465 & 0.06422 \\ \cline{2-7} 
 & MARBERTv02 & 0.85727 & 0.8617 & 0.86469 & 0.84997 & 0.0647 \\ \cline{2-7} 
 & DeBERTa-v3-base & \textbf{0.9406} & \textbf{0.94154} & \textbf{0.94216} & \textbf{0.93905} & \textbf{0.02711} \\ \hline
 QALB-15 L1 Test Set & AraBERTv02 & 0.91782 & 0.92959 & 0.9376 & 0.89886 & 0.03423 \\ \cline{2-7} 
 & CAMeLBERT-MSA & 0.90663 & 0.9216 & 0.93187 & 0.88272 & 0.03867 \\ \cline{2-7} 
 & ARBERTv02 & 0.86732 & 0.8779 & 0.88509 & 0.85026 & 0.0553 \\ \cline{2-7} 
 & MARBERTv02 & 0.85723 & 0.85942 & 0.86089 & 0.8536 & 0.06048 \\ \cline{2-7} 
 & DeBERTa-v3-base & \textbf{0.94834} & \textbf{0.95282} & \textbf{0.95583} & \textbf{0.94097} & \textbf{0.0218} \\ \hline
 QALB-15 L2 Test Set & AraBERTv02 & 0.81139 & 0.81068 & 0.8102 & 0.81259 & 0.18695 \\ \cline{2-7} 
 & CAMeLBERT-MSA & 0.8002 & 0.8173 & 0.74039 & 0.77324 & 0.19109 \\ \cline{2-7} 
 & ARBERTv02 & 0.79836 & 0.78812 & 0.78144 & 0.81604 & 0.20399 \\ \cline{2-7} 
 & MARBERTv02 & 0.80123 & 0.78239 & 0.77031 & 0.83473 & 0.20497 \\ \cline{2-7} 
 & DeBERTa-v3-base & \textbf{0.81628} & \textbf{0.81761} & \textbf{0.8185} & \textbf{0.81407} & \textbf{0.18135} \\ \hline
\end{tabular}%
}
\end{table*}

According to Table 5, Arabic language models perform differently when applied to the Dev Set derived from QALB-14-L1 and QALB-15-L1 data. 
The best-performing model among all models was DeBERTa-v3-base, which scored an F1 value of 0.90834. The superior performance of DeBERTa is the result of its ability to leverage advanced architecture that relies on attention mechanisms more effectively than other models. Arabic multi-label classification requires understanding the relationships between words in a sentence because of its morphological and syntactic complexity. This is an advantage offered by DeBERTa's innovations. DeBERTa also uses an enhanced mask decoder to analyze pre-training tasks and adapt them to the model's scale for better performance on subsequent tasks; this could enhance the model's classification of multi-label Arabic classification. DeBERTa represents a word using two vectors: a first vector embeds the word, and a second vector embeds its position. The attention weights of words are then calculated by detangling matrices of relative position and content. With BERT, only one vector is used to represent the sum of word embedding and position embedding for a given word. A critical component of DeBERTa's training is Masked Language Modeling (MLM), which involves learning by observing the relationship between word content and position in context. The performance of DeBERTa V3~\cite{he2021debertav3} has been improved through the replacement of MLM with Replaced Token Detection (RTD).

Although AraBERTv02 was specifically designed for Arabic, it was rated 2nd after DeBERTa due to its high efficiency but faced some challenges when handling different datasets. CAMeLBERT-MSA rated 3rd with 0.88519 F1 score after AraBERTv02 and DeBERTa. ARBERTv02 and MARBERTv02 performed similarly but were ranked lower than other models. DBecauseARBERTv02 was trained only on MSA, while MARBERTv02 was also trained on dialects, ARBERTv02 slightly outperformed MARBERTv02. 

Even though Arabic text-specific models like MARBERTv02 are designed specifically for analyzing Arabic texts, they perform worse than cross-linguistic models such as DeBERTav3-base, which are made to multilingual. In terms of dealing with different dialects and styles of Arabic, cross-linguistic models could benefit from larger training data and better generalization.

The AraBERTv02 and the DeBERTa-v3-base models both demonstrate high levels of precision compared to other models, with AraBERTv02 scoring 8.909 and DeBERTa-v3-base scoring 8.91673. Both models are capable of making accurate predictions without generating a great deal of false positives. DeBERTa-v3-base is capable of capturing the majority of true positives in the data, according to its recall statistic of 0.9001. Comparatively, MARBERTv02 has a lower recall (0.83213), meaning there are some undetected positives.

According to the Hamming Loss value for DeBERTa-v3-base, it had the lowest error rate in label classification (0.04802). Comparatively, ARBERTv02 and MARBERTv02  showed higher error rates (0.08109 and 0.07393, respectively). QALB texts contain spelling or contextual errors, which may explain the differing performance of these models compared to the data they were trained on.

Table 6 shows the evaluation result combining the test data for QALB-14 and QALB-15. For all models, the "Test Set" was more demanding compared to the development set, illustrating the model's generalizability. In the table below, test results are shown that are relatively similar to development results. DeBERTa-v3-base outperformed all other models with an F1 score of 0.94417, demonstrating its ability to handle unfamiliar data. In addition to DeBERTa, AraBERTv0,2, and CAMeLBERT-MSA also showed strong performance, with AraBERTv02 recording an F1 value of 0.91328 and CAMeLBERT-MSA recording an F1 value of 0.90554, demonstrating their ability to adapt to new texts.

AarBERTv02 and CameLBERT-MSA, which were specifically trained for Arabic, perform competitively, although further improvements may be necessary to exceed the performance of the f DeBERTa-v3-base, which is a state-of-the-art cross-linguistic model. Taking advantage of recent advances in language model architecture is important based on the high performance of DeBERTa. According to the analysis, precision and recall must be balanced, and optimal thresholds must be set to achieve the desired performance in practical applications, such as language correction and grammatical error correction.

\begin{table*}[t]
\centering
\setlength{\tabcolsep}{5pt} 
\renewcommand{\arraystretch}{1.5} 
\caption{A comparison of different BERT models based on evaluation metrics on combined QALB-14 and  QALB-15-L1 dev sets.}

\resizebox{\textwidth}{!}{%
\begin{tabular}{|m{2.5cm}|l|l|l|l|l|l|l|}
\hline
\textbf{Data} &
  \textbf{Model} &
  \textbf{F1} &
  \textbf{F0.5} &
  \textbf{Precision} &
  \textbf{Recall} &
  \textbf{Hamming Loss} &
  \textbf{Best F1 threshold} \\ \hline
QALB-14-L1 QALB-15-L1 Dev set &
  AraBERTv02 &
  0.89401 &
  0.90406 &
  0.9109 &
  0.87773 &
  0.05446 &
  0.831 \\ \cline{2-8} 
 & CAMeLBERT-MSA & 0.88519 & 0.89494 & 0.90156 & 0.86941 & 0.05965 & 0.12 \\ \cline{2-8} 
 & ARBERTv02     & 0.85801 & 0.86597 & 0.87136 & 0.84505 & 0.07393 & 0.56 \\ \cline{2-8} 
 & MARBERTv02    & 0.84435 & 0.85186 & 0.85694 & 0.83213 & 0.08109 & 0.4  \\ \cline{2-8} 
 &
  DeBERTa-v3-base &
  \textbf{0.90834} &
  \textbf{0.91335} &
  \textbf{0.91673} &
  \textbf{0.9001} &
  \textbf{0.04802} &
  0.52 \\ \hline
\end{tabular}%
}
\end{table*}

\begin{table*}[t]
\centering
\setlength{\tabcolsep}{5pt} 
\renewcommand{\arraystretch}{1.5} 
\caption{A comparison of different BERT models based on evaluation metrics on combined QALB-14 and  QALB-15-L1 Test sets.}

\resizebox{\textwidth}{!}{%
\begin{tabular}{|m{2.5cm}|l|l|l|l|l|l|l|}
\hline
  \textbf{Data} &
  \textbf{Model} &
  \textbf{F1} &
  \textbf{F0.5} &
  \textbf{Precision} &
  \textbf{Recall} &
  \textbf{Hamming Loss} &
  \textbf{Best F1 threshold} \\ \hline
 QALB-14-L1 QALB-15-L1 Test set &
  AraBERTv02 &
  0.91392 &
  0.92623 &
  0.93463 &
  0.89411 &
  0.0372 &
  0.48 \\ \cline{2-8} 
 & CAMeLBERT-MSA & 0.90554 & 0.92081 & 0.93127 & 0.88119 & 0.0406  & 0.5  \\ \cline{2-8} 
 & ARBERTv02     & 0.86205 & 0.87112 & 0.87964 & 0.84734 & 0.05989 & 0.57 \\ \cline{2-8} 
 & MARBERTv02    & 0.85717 & 0.86203 & 0.86531 & 0.84918 & 0.0625  & 0.37 \\ \cline{2-8} 
 &
  DeBERTa-v3-base &
  \textbf{0.94417} &
  \textbf{0.94677} &
  \textbf{0.94852} &
  \textbf{0.93986} &
  \textbf{0.02455} &
  \textbf{0.64} \\ \hline
\end{tabular}%
}
\end{table*}
\subsection{Impact of DeBERTav3 for Predicting each Error Tag}
A detailed evaluation of the model's performance is provided in Table 7 using four key metrics: precision, recall, F1 score, and F0.5 score. Using QALB-14 and QALB-15 data, we evaluated the Dev Set and the test set.  Several categories, including OH, OT, OA, PC, and PM, demonstrated excellent performance, achieving high precision, recall, and F1 values in both the development and test sets. For example, the OH class achieved an F1 score of 0.9817 in the development set and 0.9925 in the test set, which indicates the model was accurate at recognizing it. In the test set, the PM class reached almost perfect results (F1 score = 1.000), which indicates consistent performance and low error rates. However, some classes (such as ON and OS) achieved zero precision, recall, or F1 score, due to limited samples available (support = 1 and 8 in the development set, respectively). Due to limited data, the model had difficulty handling these classes. MT demonstrated unstable performance in both sets (development and test ), reaching F1 scores of 0.367 in development, but 0.273 in the test, to occur even with a high precision of 0.75. This indicates a poor recall and most samples failed to be detected. Moreover, in some classes, test and development sets showed variations in performance. OG, for example, achieved an F1 score of 0.738 in the development set, but increased to 0.865 in the test set, indicating improved generalization. On the other hand, the OC class showed unstable performance, since the F1 score for this class was 0.500 in the development set, and 0.548 in the test set, indicating difficulty dealing with this class.

Model performance was shown to be stable when all samples were treated equally, with a Micro-Average of 0.908 in the development set and 0.944 in the test set. The mode can handle dominant classes efficiently, resulting in this improvement. Compared to the micro-average, the macro-average was lower (0.775 in development, 0.804 in test), indicating a decrease in average performance due to some low-performing classes. Weighted averages (F1 = 0.908 in development, 0.943 in testing) were close to micro-averages, reflecting the influence of highly supported classes (such as OH and PM).

The results showed that the model performed well on most high-frequency classes (e.g., OH and PM), but struggled with low-frequency classes (e.g., ON and OS). Despite the strong performance of the micro and weighted measures, the low macro rates indicate that underperforming classes need to be handled better. As a result of future improvements, the model should be able to generalize across classes and address data imbalance.
\begin{table*}[t]
\centering
\setlength{\tabcolsep}{5pt} 
\renewcommand{\arraystretch}{1.5} 
\caption{Impact of DeBERTav3 model in predicting each error tag separately.}

\resizebox{\textwidth}{!}{%
\begin{tabular}{l|lllll|lllll}
\cline{2-11}
 &
  \multicolumn{5}{l|}{\textbf{QALB-14 and QALB-15 Dev Set}} &
  \multicolumn{5}{l|}{\textbf{QALB-14 and QALB-15 Test Set}} \\ \hline
\multicolumn{1}{|l|}{\textbf{Class}} &
  \multicolumn{1}{l|}{\textbf{Precision}} &
  \multicolumn{1}{l|}{\textbf{Recall}} &
  \multicolumn{1}{l|}{\textbf{F1 Score}} &
  \multicolumn{1}{l|}{\textbf{F0.5 Score}} &
  \textbf{Support} &
  \multicolumn{1}{l|}{\textbf{Precision}} &
  \multicolumn{1}{l|}{\textbf{Recall}} &
  \multicolumn{1}{l|}{\textbf{F1 Score}} &
  \multicolumn{1}{l|}{\textbf{F0.5 Score}} &
  \multicolumn{1}{l|}{\textbf{Support}} \\ \hline
\multicolumn{1}{|l|}{\textbf{OH}} &
  \multicolumn{1}{l|}{0.989076} &
  \multicolumn{1}{l|}{0.974560} &
  \multicolumn{1}{l|}{0.981764} &
  \multicolumn{1}{l|}{0.986139} &
  1022.0 &
  \multicolumn{1}{l|}{0.995160} &
  \multicolumn{1}{l|}{0.989771} &
  \multicolumn{1}{l|}{0.992459} &
  \multicolumn{1}{l|}{0.994078} &
  \multicolumn{1}{l|}{1662.0} \\ \hline
\multicolumn{1}{|l|}{\textbf{OT}} &
  \multicolumn{1}{l|}{0.986063} &
  \multicolumn{1}{l|}{0.965870} &
  \multicolumn{1}{l|}{0.975862} &
  \multicolumn{1}{l|}{0.981957} &
  293.0 &
  \multicolumn{1}{l|}{0.989712} &
  \multicolumn{1}{l|}{0.995859} &
  \multicolumn{1}{l|}{0.992776} &
  \multicolumn{1}{l|}{0.990935} &
  \multicolumn{1}{l|}{483.0} \\ \hline
\multicolumn{1}{|l|}{\textbf{OA}} &
  \multicolumn{1}{l|}{0.954545} &
  \multicolumn{1}{l|}{0.940299} &
  \multicolumn{1}{l|}{0.947368} &
  \multicolumn{1}{l|}{0.951662} &
  201.0 &
  \multicolumn{1}{l|}{0.983099} &
  \multicolumn{1}{l|}{0.991477} &
  \multicolumn{1}{l|}{0.987270} &
  \multicolumn{1}{l|}{0.984763} &
  \multicolumn{1}{l|}{352.0} \\ \hline
\multicolumn{1}{|l|}{\textbf{OW}} &
  \multicolumn{1}{l|}{0.978723} &
  \multicolumn{1}{l|}{0.821429} &
  \multicolumn{1}{l|}{0.893204} &
  \multicolumn{1}{l|}{0.942623} &
  112.0 &
  \multicolumn{1}{l|}{0.976744} &
  \multicolumn{1}{l|}{0.893617} &
  \multicolumn{1}{l|}{0.933333} &
  \multicolumn{1}{l|}{0.958904} &
  \multicolumn{1}{l|}{141.0} \\ \hline
\multicolumn{1}{|l|}{\textbf{ON}} &
  \multicolumn{1}{l|}{0.000000} &
  \multicolumn{1}{l|}{0.000000} &
  \multicolumn{1}{l|}{0.000000} &
  \multicolumn{1}{l|}{0.000000} &
  1.0 &
  \multicolumn{1}{l|}{0.000000} &
  \multicolumn{1}{l|}{0.000000} &
  \multicolumn{1}{l|}{0.000000} &
  \multicolumn{1}{l|}{0.000000} &
  \multicolumn{1}{l|}{0.0} \\ \hline
\multicolumn{1}{|l|}{\textbf{OS}} &
  \multicolumn{1}{l|}{0.000000} &
  \multicolumn{1}{l|}{0.000000} &
  \multicolumn{1}{l|}{0.000000} &
  \multicolumn{1}{l|}{0.000000} &
  8.0 &
  \multicolumn{1}{l|}{0.000000} &
  \multicolumn{1}{l|}{0.000000} &
  \multicolumn{1}{l|}{0.000000} &
  \multicolumn{1}{l|}{0.000000} &
  \multicolumn{1}{l|}{11.0} \\ \hline
\multicolumn{1}{|l|}{\textbf{OG}} &
  \multicolumn{1}{l|}{0.885714} &
  \multicolumn{1}{l|}{0.632653} &
  \multicolumn{1}{l|}{0.738095} &
  \multicolumn{1}{l|}{0.820106} &
  49.0 &
  \multicolumn{1}{l|}{0.872727} &
  \multicolumn{1}{l|}{0.857143} &
  \multicolumn{1}{l|}{0.864865} &
  \multicolumn{1}{l|}{0.869565} &
  \multicolumn{1}{l|}{56.0} \\ \hline
\multicolumn{1}{|l|}{\textbf{OC}} &
  \multicolumn{1}{l|}{0.511628} &
  \multicolumn{1}{l|}{0.488889} &
  \multicolumn{1}{l|}{0.500000} &
  \multicolumn{1}{l|}{0.506912} &
  45.0 &
  \multicolumn{1}{l|}{0.487805} &
  \multicolumn{1}{l|}{0.625000} &
  \multicolumn{1}{l|}{0.547945} &
  \multicolumn{1}{l|}{0.510204} &
  \multicolumn{1}{l|}{32.0} \\ \hline
\multicolumn{1}{|l|}{\textbf{OR}} &
  \multicolumn{1}{l|}{0.891026} &
  \multicolumn{1}{l|}{0.894850} &
  \multicolumn{1}{l|}{0.892934} &
  \multicolumn{1}{l|}{0.891788} &
  466.0 &
  \multicolumn{1}{l|}{0.883387} &
  \multicolumn{1}{l|}{0.861371} &
  \multicolumn{1}{l|}{0.872240} &
  \multicolumn{1}{l|}{0.878894} &
  \multicolumn{1}{l|}{642.0} \\ \hline
\multicolumn{1}{|l|}{\textbf{OD}} &
  \multicolumn{1}{l|}{0.839335} &
  \multicolumn{1}{l|}{0.846369} &
  \multicolumn{1}{l|}{0.842837} &
  \multicolumn{1}{l|}{0.840733} &
  358.0 &
  \multicolumn{1}{l|}{0.909091} &
  \multicolumn{1}{l|}{0.880361} &
  \multicolumn{1}{l|}{0.894495} &
  \multicolumn{1}{l|}{0.903196} &
  \multicolumn{1}{l|}{443.0} \\ \hline
\multicolumn{1}{|l|}{\textbf{OM}} &
  \multicolumn{1}{l|}{0.896208} &
  \multicolumn{1}{l|}{0.903421} &
  \multicolumn{1}{l|}{0.899800} &
  \multicolumn{1}{l|}{0.897641} &
  497.0 &
  \multicolumn{1}{l|}{0.852327} &
  \multicolumn{1}{l|}{0.906143} &
  \multicolumn{1}{l|}{0.878412} &
  \multicolumn{1}{l|}{0.862573} &
  \multicolumn{1}{l|}{586.0} \\ \hline
\multicolumn{1}{|l|}{\textbf{MI}} &
  \multicolumn{1}{l|}{0.921717} &
  \multicolumn{1}{l|}{0.854801} &
  \multicolumn{1}{l|}{0.886999} &
  \multicolumn{1}{l|}{0.907509} &
  427.0 &
  \multicolumn{1}{l|}{0.919694} &
  \multicolumn{1}{l|}{0.887454} &
  \multicolumn{1}{l|}{0.903286} &
  \multicolumn{1}{l|}{0.913060} &
  \multicolumn{1}{l|}{542.0} \\ \hline
\multicolumn{1}{|l|}{\textbf{MT}} &
  \multicolumn{1}{l|}{0.384615} &
  \multicolumn{1}{l|}{0.350877} &
  \multicolumn{1}{l|}{0.366972} &
  \multicolumn{1}{l|}{0.377358} &
  57.0 &
  \multicolumn{1}{l|}{0.750000} &
  \multicolumn{1}{l|}{0.166667} &
  \multicolumn{1}{l|}{0.272727} &
  \multicolumn{1}{l|}{0.441176} &
  \multicolumn{1}{l|}{18.0} \\ \hline
\multicolumn{1}{|l|}{\textbf{XC}} &
  \multicolumn{1}{l|}{0.924242} &
  \multicolumn{1}{l|}{0.905941} &
  \multicolumn{1}{l|}{0.915000} &
  \multicolumn{1}{l|}{0.920523} &
  404.0 &
  \multicolumn{1}{l|}{0.934028} &
  \multicolumn{1}{l|}{0.957295} &
  \multicolumn{1}{l|}{0.945518} &
  \multicolumn{1}{l|}{0.938590} &
  \multicolumn{1}{l|}{562.0} \\ \hline
\multicolumn{1}{|l|}{\textbf{XF}} &
  \multicolumn{1}{l|}{0.900585} &
  \multicolumn{1}{l|}{0.900585} &
  \multicolumn{1}{l|}{0.900585} &
  \multicolumn{1}{l|}{0.900585} &
  171.0 &
  \multicolumn{1}{l|}{0.843750} &
  \multicolumn{1}{l|}{0.915254} &
  \multicolumn{1}{l|}{0.878049} &
  \multicolumn{1}{l|}{0.857143} &
  \multicolumn{1}{l|}{59.0} \\ \hline
\multicolumn{1}{|l|}{\textbf{XG}} &
  \multicolumn{1}{l|}{0.830409} &
  \multicolumn{1}{l|}{0.806818} &
  \multicolumn{1}{l|}{0.818444} &
  \multicolumn{1}{l|}{0.825581} &
  176.0 &
  \multicolumn{1}{l|}{0.786325} &
  \multicolumn{1}{l|}{0.779661} &
  \multicolumn{1}{l|}{0.782979} &
  \multicolumn{1}{l|}{0.784983} &
  \multicolumn{1}{l|}{118.0} \\ \hline
\multicolumn{1}{|l|}{\textbf{XN}} &
  \multicolumn{1}{l|}{0.670391} &
  \multicolumn{1}{l|}{0.745342} &
  \multicolumn{1}{l|}{0.705882} &
  \multicolumn{1}{l|}{0.684151} &
  161.0 &
  \multicolumn{1}{l|}{0.860000} &
  \multicolumn{1}{l|}{0.745665} &
  \multicolumn{1}{l|}{0.798762} &
  \multicolumn{1}{l|}{0.834411} &
  \multicolumn{1}{l|}{173.0} \\ \hline
\multicolumn{1}{|l|}{\textbf{XT}} &
  \multicolumn{1}{l|}{0.950588} &
  \multicolumn{1}{l|}{0.943925} &
  \multicolumn{1}{l|}{0.947245} &
  \multicolumn{1}{l|}{0.949248} &
  428.0 &
  \multicolumn{1}{l|}{0.945493} &
  \multicolumn{1}{l|}{0.933747} &
  \multicolumn{1}{l|}{0.939583} &
  \multicolumn{1}{l|}{0.943120} &
  \multicolumn{1}{l|}{483.0} \\ \hline
\multicolumn{1}{|l|}{\textbf{XM}} &
  \multicolumn{1}{l|}{0.870536} &
  \multicolumn{1}{l|}{0.752896} &
  \multicolumn{1}{l|}{0.807453} &
  \multicolumn{1}{l|}{0.844156} &
  259.0 &
  \multicolumn{1}{l|}{0.759740} &
  \multicolumn{1}{l|}{0.632432} &
  \multicolumn{1}{l|}{0.690265} &
  \multicolumn{1}{l|}{0.730337} &
  \multicolumn{1}{l|}{185.0} \\ \hline
\multicolumn{1}{|l|}{\textbf{SW}} &
  \multicolumn{1}{l|}{0.881356} &
  \multicolumn{1}{l|}{0.864266} &
  \multicolumn{1}{l|}{0.872727} &
  \multicolumn{1}{l|}{0.877884} &
  361.0 &
  \multicolumn{1}{l|}{0.904891} &
  \multicolumn{1}{l|}{0.822222} &
  \multicolumn{1}{l|}{0.861578} &
  \multicolumn{1}{l|}{0.887054} &
  \multicolumn{1}{l|}{405.0} \\ \hline
\multicolumn{1}{|l|}{\textbf{SF}} &
  \multicolumn{1}{l|}{0.818605} &
  \multicolumn{1}{l|}{0.846154} &
  \multicolumn{1}{l|}{0.832151} &
  \multicolumn{1}{l|}{0.823970} &
  208.0 &
  \multicolumn{1}{l|}{0.883534} &
  \multicolumn{1}{l|}{0.948276} &
  \multicolumn{1}{l|}{0.914761} &
  \multicolumn{1}{l|}{0.895765} &
  \multicolumn{1}{l|}{232.0} \\ \hline
\multicolumn{1}{|l|}{\textbf{PC}} &
  \multicolumn{1}{l|}{0.983908} &
  \multicolumn{1}{l|}{0.963964} &
  \multicolumn{1}{l|}{0.973834} &
  \multicolumn{1}{l|}{0.979853} &
  444.0 &
  \multicolumn{1}{l|}{0.993590} &
  \multicolumn{1}{l|}{0.982567} &
  \multicolumn{1}{l|}{0.988048} &
  \multicolumn{1}{l|}{0.991366} &
  \multicolumn{1}{l|}{631.0} \\ \hline
\multicolumn{1}{|l|}{\textbf{PT}} &
  \multicolumn{1}{l|}{0.524862} &
  \multicolumn{1}{l|}{0.904762} &
  \multicolumn{1}{l|}{0.664336} &
  \multicolumn{1}{l|}{0.572979} &
  105.0 &
  \multicolumn{1}{l|}{0.977444} &
  \multicolumn{1}{l|}{0.992366} &
  \multicolumn{1}{l|}{0.984848} &
  \multicolumn{1}{l|}{0.980392} &
  \multicolumn{1}{l|}{262.0} \\ \hline
\multicolumn{1}{|l|}{\textbf{PM}} &
  \multicolumn{1}{l|}{1.000000} &
  \multicolumn{1}{l|}{0.996422} &
  \multicolumn{1}{l|}{0.998208} &
  \multicolumn{1}{l|}{0.999282} &
  1118.0 &
  \multicolumn{1}{l|}{1.000000} &
  \multicolumn{1}{l|}{1.000000} &
  \multicolumn{1}{l|}{1.000000} &
  \multicolumn{1}{l|}{1.000000} &
  \multicolumn{1}{l|}{1745.0} \\ \hline
\multicolumn{1}{|l|}{\textbf{MG}} &
  \multicolumn{1}{l|}{0.939227} &
  \multicolumn{1}{l|}{0.949721} &
  \multicolumn{1}{l|}{0.944444} &
  \multicolumn{1}{l|}{0.941307} &
  358.0 &
  \multicolumn{1}{l|}{0.962617} &
  \multicolumn{1}{l|}{0.979398} &
  \multicolumn{1}{l|}{0.970935} &
  \multicolumn{1}{l|}{0.965927} &
  \multicolumn{1}{l|}{631.0} \\ \hline
\multicolumn{1}{|l|}{\textbf{SP}} &
  \multicolumn{1}{l|}{1.000000} &
  \multicolumn{1}{l|}{0.724138} &
  \multicolumn{1}{l|}{0.840000} &
  \multicolumn{1}{l|}{0.929204} &
  319.0 &
  \multicolumn{1}{l|}{0.997409} &
  \multicolumn{1}{l|}{0.994832} &
  \multicolumn{1}{l|}{0.996119} &
  \multicolumn{1}{l|}{0.996893} &
  \multicolumn{1}{l|}{387.0} \\ \hline
\multicolumn{1}{|l|}{\textbf{Average Metrics:}} &
  \multicolumn{4}{c|}{QALB-14 and QALB-15 Dev Set} &
   &
  \multicolumn{4}{c|}{QALB-14 and QALB-15 Test Set} &
   \\ \cline{1-5} \cline{7-10}
\multicolumn{1}{|l|}{\textit{\textbf{\begin{tabular}[c]{@{}l@{}}Average \\ Type\end{tabular}}}} &
  \multicolumn{1}{l|}{\textit{\textbf{Precision}}} &
  \multicolumn{1}{l|}{\textit{\textbf{Recall}}} &
  \multicolumn{1}{l|}{\textit{\textbf{F1 Score}}} &
  \multicolumn{1}{l|}{\textit{\textbf{F0.5 score}}} &
  \textit{\textbf{}} &
  \multicolumn{1}{l|}{\textit{\textbf{Precision}}} &
  \multicolumn{1}{l|}{\textit{\textbf{Recall}}} &
  \multicolumn{1}{l|}{\textit{\textbf{F1 Score}}} &
  \multicolumn{1}{l|}{\textit{\textbf{F0.5 Score}}} &
  \textit{\textbf{}} \\ \cline{1-5} \cline{7-10}
\multicolumn{1}{|l|}{\textit{Micro}} &
  \multicolumn{1}{l|}{0.916730} &
  \multicolumn{1}{l|}{0.900099} &
  \multicolumn{1}{l|}{0.908339} &
  \multicolumn{1}{l|}{0.913355} &
   &
  \multicolumn{1}{l|}{0.948520} &
  \multicolumn{1}{l|}{0.939858} &
  \multicolumn{1}{l|}{0.944169} &
  \multicolumn{1}{l|}{0.946775} &
   \\ \cline{1-5} \cline{7-10}
\multicolumn{1}{|l|}{\textit{Macro}} &
  \multicolumn{1}{l|}{0.789745} &
  \multicolumn{1}{l|}{0.768421} &
  \multicolumn{1}{l|}{0.774852} &
  \multicolumn{1}{l|}{0.782813} &
   &
  \multicolumn{1}{l|}{0.825714} &
  \multicolumn{1}{l|}{0.797638} &
  \multicolumn{1}{l|}{0.803510} &
  \multicolumn{1}{l|}{0.812051} &
   \\ \cline{1-5} \cline{7-10}
\multicolumn{1}{|l|}{\textit{Weighted}} &
  \multicolumn{1}{l|}{0.920939} &
  \multicolumn{1}{l|}{0.900099} &
  \multicolumn{1}{l|}{0.908471} &
  \multicolumn{1}{l|}{0.915442} &
   &
  \multicolumn{1}{l|}{0.947141} &
  \multicolumn{1}{l|}{0.939858} &
  \multicolumn{1}{l|}{0.942776} &
  \multicolumn{1}{l|}{0.945094} &
   \\ \cline{1-5} \cline{7-10}
\end{tabular}%
}
\end{table*}
\subsection{Impact of Synthetic Data on GEC Task}

To evaluate the generated synthetic data, we compared them with Solyman et al.~\cite{solyman2023optimizing} considered the last research to generate Arabic synthetic data. We trained a grammatical error correction model based on the data generated by Solyman et al.~\cite{solyman2023optimizing} using AraT5, and a second model using the synthetic data generated using our model. To generate our data, we only extracted correct data from Solyman et al.~\cite{solyman2023optimizing}, then we fed the correct sentences to the error tagging model, which produced the possible errors, after that, fed the output of the error tagging model which is predicting tags and correct sentence to our synthetic data generation model that produced the incorrect sentences parallel to the correct ones. Our goal is to evaluate the same synthetic data generated by~\cite{solyman2023optimizing} with the data produced by our model on the same grammatical error correction model.

\begin{table*}[t]
\centering
\caption{A GEC model was trained using only synthetic data with AraT5.}

\resizebox{\textwidth}{!}{%
\begin{tabular}{l|ccccc|ccccc|}
\cline{2-11}
 & \multicolumn{5}{c|}{\textbf{QALB-14}} & \multicolumn{5}{c|}{\textbf{QALB-15}} \\ \cline{2-11} 
 & \multicolumn{1}{c|}{\textbf{Precision}} & \multicolumn{1}{c|}{\textbf{Recall}} & \multicolumn{1}{c|}{\textbf{F1}} & \multicolumn{1}{c|}{\textbf{F0.5}} & \textbf{BLUE-4} & \multicolumn{1}{c|}{\textbf{Precision}} & \multicolumn{1}{c|}{\textbf{Recall}} & \multicolumn{1}{c|}{\textbf{F1}} & \multicolumn{1}{c|}{\textbf{F0.5}} & \textbf{BLUE-4} \\ \hline
\multicolumn{1}{|l|}{\textbf{\begin{tabular}[c]{@{}l@{}}Synthetic data generation in \\ Solyman et al.~\cite{solyman2023optimizing}  +Transformer GEC\end{tabular}}} & \multicolumn{1}{c|}{73.05} & \multicolumn{1}{c|}{53.91} & \multicolumn{1}{c|}{62.02} & \multicolumn{1}{c|}{\textbf{-}} & 74.39 & \multicolumn{1}{c|}{71.94} & \multicolumn{1}{c|}{58.88} & \multicolumn{1}{c|}{64.74} & \multicolumn{1}{c|}{\textbf{-}} & 77.82 \\ \hline
\multicolumn{1}{|l|}{\textbf{\begin{tabular}[c]{@{}l@{}}Synthetic data generation in \\ Solyman et al.~\cite{solyman2023optimizing} + ARAT5 GEC\end{tabular}}} & \multicolumn{1}{c|}{81.80} & \multicolumn{1}{c|}{69.40} & \multicolumn{1}{c|}{75.09} & \multicolumn{1}{c|}{78.98} & \textbf{90.13} & \multicolumn{1}{c|}{\textbf{73.85}} & \multicolumn{1}{c|}{\textbf{58.09}} & \multicolumn{1}{c|}{\textbf{64.25}} & \multicolumn{1}{c|}{\textbf{69.41}} & 80.22 \\ \hline
\multicolumn{1}{|l|}{\textbf{\begin{tabular}[c]{@{}l@{}}Our synthetic data generation + \\ ARAT5 GEC\end{tabular}}} & \multicolumn{1}{c|}{\textbf{81.88}} & \multicolumn{1}{c|}{\textbf{70.68}} & \multicolumn{1}{c|}{\textbf{75.87}} & \multicolumn{1}{c|}{\textbf{79.36}} & 86.73 & \multicolumn{1}{c|}{73.70} & \multicolumn{1}{c|}{56.77} & \multicolumn{1}{c|}{62.99} & \multicolumn{1}{c|}{68.57} & \textbf{84.75} \\ \hline
\end{tabular}%
}
\end{table*}
According to Table 8, different models perform differently in correcting grammatical errors in syntactic data on QALB-14 and QALB-15 datasets using precision, recall, F1, and BLEU-4 metrics. Synthetic data has been shown to significantly improve the performance of the grammatical error correction model, especially when combined with the ARAT5 model.

The first model was developed using synthetic Solyman et al.~\cite{solyman2023optimizing} data only on a transformer-based model, but while the synthetic data represents an effective approach to developing a syntactic error correction model, the performance was relatively limited, with a precision value of 73.05\% on the QALB-14 test set and 71.94\% on the QALB-15 test set. Based on these results, it is evident how important it is to rely on synthetic data that covers a wide range of errors and is representative of the actual data since BLUE-4 results of 74.39 and 77.8 indicate moderate effectiveness in producing accurate corrections.

A second Solyman et al.~\cite{solyman2023optimizing} data that uses the ARAT5 GEC with synthetic data resulted in a significant improvement in indicator precision, with 81.80\% for QALB-14 and 73.85\% for QALB-15. This improvement is evidence of the ability of ARAT5 to enhance the quality of correction by improving the ability to capture more complex errors, and the significant increase in the BLUE-4 index (reaching 90.13 and 80.22) confirms the quality and accuracy of correction have improved significantly.

Synthetic data generated by our model combined with ARAT5 GEC led to the best results in this study. A QALB-14 set was 81.88\% precision, a QALB-15 set was 73.70\% precision, and a QALB-15 set was 56.77\% accurate. In the following performance test, the BLUE-4 model produced more accurate and natural language expression corrections by reaching 86.73\% and 84.75\%, respectively. These results demonstrate how synthetic data can improve the model's performance by covering various types of errors. Our model appears to have been improved by the data generated, both in terms of diversity and accuracy, for the task of correcting grammatical errors.

The use of specially designed synthetic data can significantly improve the performance of error correction models, especially in environments lacking sufficient labeled data. Consequently, the synthetic data we use is not only suitable for training but also helps build a more generalized model capable of correcting grammatical errors more effectively. The results indicate that the ARAT5 GEC model improves performance, as it has a positive effect on correction quality, which suggests that the combination of this model with synthetic data may prove to be an effective method for addressing Arabic grammatical errors. Performance indicators improved when compared to conventional methods: In the third experiment, the F1 index reached 75.87\% in the QALB-14 group and 62.99\% in the QALB-15 group, indicating the model outperforms the traditional approach based on Solyman data alone when using studied and diverse synthetic data.
\section{Discussion}
As a result of the error tagging model, it appears that the DeBERTaV3 model performed outstandingly in the multi-label classification of Arabic texts, outperforming the BERT models devoted to Arabic, despite being designed for multilingual training. Its superiority comes from its ability to absorb common linguistic patterns across different languages, giving it great flexibility when it comes to understanding complex Arabic features. In addition, the model is based on modern deep learning techniques such as the Multi-head Attention mechanism, which enables it to recognize different relationships between words based on context. Due to this mechanism, the model can recognize error patterns and classify them effectively based on their grammatical structure and general context. However, when compared to other Arabic models, DeBERTaV3 appears to be stronger due to its large and diverse data set, which includes several languages, including Arabic. Language diversity enhances the model's ability to generalize linguistic patterns across languages. While Arabic-specific models rely on limited linguistic data, they may be less accurate as a result of their inability to generalize. Due to its versatility, DeBERTaV3 may have been exposed to common linguistic patterns between Arabic and other languages, for example, grammatical and lexical patterns, which improved its understanding of Arabic.

When the model is applied to multi-label classification, the requirements are complex, as the model must be able to understand the context and distinguish between different categories present in a single text. As a result of DeBERTaV3's outstanding performance here, it is a robust model that is capable of understanding a wide variety of contexts and predicting effectively in complex environments. A multi-label classification of error types in the Arabic language has not been addressed in any studies in the scientific literature to date. Developing a model that is capable of classifying Arabic errors based on multiple labels is both a challenge and a potential opportunity because it will improve the quality of text processing and assist in assessing and correcting linguistic errors more effectively. As a result of such a classification, we could improve our understanding of common errors and their distribution, ultimately leading to more effective educational and correction tools, especially in academics and education.

Multi-label classification of Arabic texts is challenged by the imbalance in the distribution of classes, as some classes are more common than others, which makes it difficult to train the model to accurately represent less common classes. Consequently, the model tends to favor classes with high representation and is less accurate at classifying rare or uncommon classes, which reduces its overall accuracy and effectiveness in practical applications.

This challenge was overcome by Huang et al.~\cite{huang2021balancing} Distribution-Balanced Loss technique. A model aimed at achieving a balance in losses calculated by each class was used to compensate classes with low distributions during training by increasing their losses. Despite this technique improving the model's performance and mitigating some of the imbalance problems, the model still needs improvement to effectively deal with uneven distributions of classes.  Additionally, our synthetic data may be useful in the future for classifying Arabic texts more efficiently and flexibly by solving for rare classes.

Moreover, We developed an innovative methodology for generating Arabic syntitic sentences containing grammatical errors. The methodology uses correct sentences with possible error types and then directs the generation process to produce sentences with specific grammatical errors. This process is error-directed, as the error types may be repeated in a single sentence to enhance the model's ability to recognize different error patterns. In this methodology, back-translation is applied to generate errors for a specific purpose. Back-translation in the Arabic language is a unique contribution because this method has not been used to generate grammatical errors in previous literature. Data generated with this methodology resembles human errors, which can be used for grammatical error correction models This method opens up a lot of development opportunities, especially in the Arabic language environment, where precise data is lacking.

The final phase of the study involved training the AraT5 grammar correction model using synthetic data generated by our methodology. Comparing the model to state-of-the-art results on QALB-14 data, it achieved superior performance on precision, recall, F1, and F0.5 metrics. In tests on the QALB-15 data, the model performed well, but Solyman et al.~\cite{solyman2023optimizing} synthetic data outperformed it. The discrepancy can be attributed to the fact that the synthetic data we produced is characterized by a comprehensive diversity of errors, which allowed our model to handle grammar errors with greater flexibility. Data generated by Solyman et al.~\cite{solyman2023optimizing} are based on rule-based error input, such as spelling errors and common errors, which limits the diversity and coverage of these errors. Furthermore, our models were trained using native-speaker texts (L1 data), whereas the QALB-15 test data has a mix of native (L1) and non-native (L2) speakers. As a result of this diversity in our data, the model became more accurate, allowing it to correctly correct grammatical errors across a wide range of linguistic contexts and complex sentences.

\section{Conclusion}
This paper aims to generate synthetic data for use in developing Arabic grammatical error correction models. We develop a new method for generating synthetic data from monolingual corpora. We all know that news, newspaper articles, magazines, and other media regularly contain grammatically correct sentences. However, there is a lack of parallel corpora in Arabic. Our objective is to generate incorrect sentences by utilizing the existing correct sentences by developing two models: an error tagging model and a synthetic data generation model. Error tagging identifies the types of errors within correct sentences and appends them before correct sentences using the DeBERTav3 model. The synthetic data generation model feeds the output of the error tagging model to the AraT5 model to generate incorrect sentences. Both models achieved state-of-the-art results. In the future, we would like to develop a model for grammar correction based on our synthetic data.
\section*{Acknowledgements}
The Aziz Supercomputer at King Abdulaziz University was used for all experiments; therefore, we are grateful to the university's High-Performance Computing Center (HPC).

\bibliographystyle{plainnat}
\bibliography{references}
\end{document}